\documentclass[10pt,twocolumn,letterpaper]{article}
\usepackage[utf8]{inputenc}
\usepackage[T1]{fontenc}

\usepackage{cvpr}
\usepackage{times}
\usepackage{epsfig}

\usepackage{amsmath, amssymb, amsthm}
\usepackage{xcolor}
\usepackage[ruled,linesnumbered]{algorithm2e}
\usepackage{algorithmic}

\usepackage[pagebackref=true,breaklinks=true,letterpaper=true,colorlinks,bookmarks=false]{hyperref}

 \cvprfinalcopy 

\def\cvprPaperID{1721} 

\ifcvprfinal\fi

\def\R{\mathbb{R}}

\def\X{\mathbf{X}}
\def\Y{\mathbf{Y}}
\def\U{\mathbf{U}}
\def\V{\mathbf{V}}
\def\A{\mathbf{A}}
\def\Q{\mathbf{Q}}
\def\Z{\mathbf{Z}}
\def\I{\mathbf{I}}
\def\M{\mathbf{M}}
\def\C{\mathbf{C}}

\def\st{\ \ \text{s.t.} \ \ }

\def\z{\mathbf{z}}
\def\u{\mathbf{u}}
\def\v{\mathbf{v}}

\def\x{\mathbf{x}}
\def\y{\mathbf{y}}
\def\a{\mathbf{a}}
\def\m{\mathbf{m}}
\def\w{\mathbf{w}}
\def\balpha{\mathbf{\alpha}}

\newcommand{\myparagraph}[1]{\smallskip\noindent\textbf{#1.}}

\def\bmu{\boldsymbol{\mu}}
\def\balpha{\boldsymbol{\alpha}}
\def\Var{\mathbf{Var}}

\def\E{\mathbb{E}}
\newcommand{\F}{ {\rm F}}
\def\1{\mathbf{1}}
\def\0{\mathbf{0}}
\newcommand{\B}{\mathbf}
\newcommand{\Ber}{\text{Ber}}
\newcommand{\diag}[1]{{\rm diag}(#1)}

\newtheorem{theorem}{Theorem}
\newtheorem{lemma}[theorem]{Lemma}
\newtheorem{corollary}[theorem]{Corollary}
\newtheorem{definition}[theorem]{Definition}
\newtheorem{proposition}[theorem]{Proposition}

\begin{document}

\title{On the Regularization Properties of Structured Dropout}

\author{
{Ambar Pal \qquad Connor Lane \qquad René Vidal \qquad Benjamin D. Haeffele}\\
{Mathematical Institute for Data Science, Johns Hopkins University, Baltimore, MD} \\
{\tt\small \{ambar, clane, rvidal, bhaeffele\}@jhu.edu}
}

\maketitle
\thispagestyle{empty}

\begin{abstract}
Dropout and its extensions (\eg\ DropBlock and DropConnect) are popular heuristics for training neural networks, which have been shown to improve generalization performance in practice. However, a theoretical understanding of their optimization and regularization properties remains elusive. Recent work shows that in the case of single hidden-layer linear networks, Dropout is a stochastic gradient descent method for minimizing a regularized loss, and that the regularizer induces solutions that are low-rank and balanced. In this work we show that for single hidden-layer linear networks, DropBlock induces spectral $k$-support norm regularization, and promotes solutions that are low-rank and have factors with equal norm. We also show that the global minimizer for DropBlock can be computed in closed form, and that DropConnect is equivalent to Dropout. We then show that some of these results can be extended to a general class of Dropout-strategies, and, with some assumptions, to deep non-linear networks when Dropout is applied to the last layer. We verify our theoretical claims and assumptions experimentally with commonly used network architectures.
\end{abstract}
\vspace{-1.5em}
\section{Introduction}
\label{introduction}
Dropout is a widely-used heuristic for training deep neural networks (NN), which involves setting to zero the output of a random subset of hidden neurons at each training iteration. The improved generalization performance of Dropout in practice has led to many variants of dropout \cite{Gal:NIPS17,Gastaldi:Arxiv17,Ghiasi:NIPS18,Morerio:ICCV17}, \cite{Rennie:IEEESLT14,Wan:ICML13,Yamada:Arxiv18, Zolna:Arxiv17}. However, despite the popularity and improved empirical performance of Dropout-style techniques, several theoretical questions remain regarding their optimization and regularization properties, \eg: What objective function is minimized by general Dropout-style techniques? Do these techniques converge to a global minimum? Does Dropout-style regularization induce an explicit regularizer? What is the inductive bias of Dropout-style regularization?

\myparagraph{Related Work}
Recent work has considered some of these questions in the case of single-layer linear neural networks trained with the squared loss. For example, \cite{Cavazza:AISTATS18} shows that Dropout is a stochastic gradient descent (SGD) method for minimizing the following objective:
\begin{equation}\label{eq:DropoutNN}
\min_{\U,\V} \E_{\z} \left \| \Y - \frac{1}{\theta} \U \diag{\z} \V^\top \X \right\|_\F^2.
\end{equation}
Here $\X \in \R^{b \times N}$ and $\Y \in \R^{a \times N}$ denote a training set with $N$ training points, $\U \in \R^{a \times d}$ and $\V \in \R^{b \times d}$ are the output and input weight matrices, respectively, $d$ is number of hidden neurons, and $\z$ is a $d$-dimensional vector of Dropout variables whose $i$th entry $z_i \sim \Ber(\theta)$ is i.i.d. Bernoulli with parameter $\theta$. In addition, \cite{Cavazza:AISTATS18} shows that Dropout induces explicit regularization in the form of a squared nuclear norm, which is known to induce low-rank solutions. Specifically,  \cite{Cavazza:AISTATS18} shows that the optimization problem \eqref{eq:DropoutNN} reduces to
\begin{equation}
\label{eq:DropoutNN-variational}
\min_{\U,\V} \|\Y - \U\V^\top\X\|_F^2 + \frac{1-\theta}{\theta} \sum_{i=1}^d \|\u_i\|_2^2\|\X^\top \v_i\|_2^2,
\end{equation}
where $\u_i$ and $\v_i$ denote the $i^\text{th}$ columns of $\U$ and $\V$, resp.
Moreover, \cite{Cavazza:AISTATS18} shows that a global minimum $(\U^*,\V^*)$ of \eqref{eq:DropoutNN-variational} yields a global minimum of $\min_{\Z} \|\Y - \Z\|_F^2 + \lambda \|\Z\|_*^2$, where $\Z^* = \U^*\V^{*\top}\X$ and $\|\Z\|_*$ is the nuclear norm. In addition, \cite{Mianjy:ICML18} shows that the optimal weights $(\U^*,\V^*)$ can be found in polynomial time and are balanced, i.e., the product of the norms of incoming and outgoing weights, $\|\u_i\|_2\|\X^\top \v_i\|_2$, is the same for all neurons. 

\myparagraph{Paper Contributions}
In this paper, we significantly generalize these results to more general Dropout schemes and more general neural network architectures.
We first study DropBlock, an alternative to Dropout for convolutional networks which was recently proposed in \cite{Ghiasi:NIPS18} and displays improved performance compared to Dropout in practice. Instead of zeroing the output of each neuron independently, DropBlock introduces a structural dropping pattern by zeroing a block of neurons within a local neighborhood together to reflect the strong correlations in responses for neighboring pixels in a CNN. Specifically, for a block-size $r$, we will look at the following optimization problem:
\begin{equation}\label{eq:DropBlockNN}
\min_{\U,\V} \E_{\w} \left \| \Y - \frac{1}{\theta} \U (\diag{\w} \otimes \I_{\frac{d}{r}}) \V^\top \X \right\|_\F^2,
\end{equation}
where $\otimes$ denotes the Kronecker product and $\w$ are the stochastic Bernoulli variables with one entry $w_i \sim {\rm Ber}(\theta)$ getting applied simultaneously to a block of columns in $(\U,\V)$ of size $r$.
In this paper, we study the regularization properties of DropBlock, and show that it induces low-rank regularization in the form of a $k$-support norm on the singular values of the solution, which is known to have some favorable properties compared to the $\ell_1$-norm \cite{Argyriou:NIPS12}. This provides a step towards explaining the superior performance of DropBlock in practice, as compared to Dropout, which induces an $\ell_1$-norm on the singular values.
In this paper we also study the properties of the optimal solutions induced by DropBlock. Specifically, we prove that the solutions to \eqref{eq:DropBlockNN} are such that the norms of the \emph{factors} are balanced, \ie\ products of corresponding blocks of $r$ columns of $\U$ and $\V$ have equal Frobenius norms. Combining these results will allow us to get a closed form solution to~\eqref{eq:DropBlockNN}. 

We then extend our analysis to more general dropping strategies that allow for arbitrary sampling distributions for the Dropout variables and obtain the explicit regularizer for this general case. We also extend our analysis to Dropout applied to the last layer of a 
deep neural network and show that this as well as many existing results in the literature can be readily extended to this scenario. We end with a short result on an equivalence between Dropout and DropConnect, which is a different way of performing Dropout. 

Finally, various experiments are used to validate the theoretical results and assumptions whenever necessary.

\section{DropBlock Analysis}
\label{DropBlock}


In this section, we study the optimization and regularization properties of DropBlock, a variant of Dropout where blocks of neurons are dropped together. In this setting, we let $d$ be the final hidden layer dimension and let $r$ be the size of the block that is dropped. We make the simplifying assumption that the blocks form a partition of the neurons in the final hidden layer, which requires the hidden dimension $d$ to be a multiple of $r$. This is a minor assumption when $d \gg r$, which is typically satisfied.\footnote{We show experimentally in Section \ref{approxverify} that a scaling of $\theta$ suffices to make this approximation behave largely identical to the original DropBlock strategy.}  Then at each iteration, we sample a binary vector of $k = \tfrac{d}{r}$ i.i.d. $\Ber(\theta)$ random variables $\w \in \{0,1\}^k$ and set the corresponding block of variables in $\z \in\{0,1\}^d$ to the value of $w_i$, i.e., $z_j = w_i$ for $(i-1)r  < j \leq ir$. This sampling scheme, which we refer to as \texttt{DropBlockSample}$(\theta, r)$, captures the key principle behind DropBlock by dropping a block of neighboring neurons at a time and is a very close approximation of DropBlock (which does not assume the blocks need to be non-overlapping) when $d \gg r$. 
The resulting DropBlock algorithm that we will study is specified in Algorithm~\ref{alg:sgd}.
Note that Dropout can be obtained as a particular case of DropBlock when the block size is set to $r = 1$. 

\myparagraph{Analysis Technique} Before presenting the details of our analysis in the subsequent subsections, we pause and comment on the analysis approach at a high level. Our goal is to understand the regularization induced by DropBlock training, i.e. Algorithm \ref{alg:sgd}. We begin in Section \ref{induced-reg} by observing that DropBlock training is equivalent to training the original un-regularized network with an additional regularization term. We continue in Section \ref{capacity-control} by analyzing what happens to this regularizer when the network width is allowed to grow arbitrarily, and observe that the regularizer value goes to $0$, hence providing no regularization to the problem. To address this issue, in Sec \ref{balanced-weights} we adaptively scale the dropout rate as a function of the network width and show that this induces a modified optimization problem whose optimal weights are balanced. These results are used in Sec \ref{k-support} to obtain a convex
lower bound to the optimization objective, which is shown to be tight, hence allowing us to relate the solutions of the convex lower bound to the solutions of the original objective of interest. Finally, we obtain a closed form solution to the convex lower bound, which additionally allows us to characterize solutions to the non-convex DropBlock optimization problem in closed form.
%

\setlength{\textfloatsep}{8pt}
\begin{algorithm}[t]
\caption{\label{alg:sgd} Dropblock Algorithm}
\begin{algorithmic}[1]
\STATE \textbf{Input:} Training Data $\mathcal{D} = \{\x_t, \y_t\}$, Learning Rate $\eta$, Retain Probability $\theta$, Block Size $r$
\STATE \textbf{Output:} Final Iterates $\U_T, \V_T$
\STATE $\U_0 \gets \U_{\rm init}$, $\V_0 \gets \V_{\rm init}$
\FOR{$t = 1, \ldots, T$}
\STATE $\z_{t-1} \gets \texttt{DropBlockSample}(\theta, r)$  \label{samplingstep}
\STATE  $\B{D}_\z \gets \diag{\z_{t-1}}$
\STATE Error, $\epsilon_t \gets \left( (\frac{1}{\theta} \U_{t - 1} \B{D}_\z \V_{t - 1}^\top \x_t) - \y_t \right)$
\STATE $\U_{t} \gets \U_{t - 1} - \ \frac{\eta}{\theta}\epsilon_t \x_t^\top \V_{t - 1} \B{D}_\z$ \label{step:GradU}
\STATE $\V_{t} \gets \V_{t - 1} - \ \frac{\eta}{\theta}\x_t \epsilon_t^\top \U_{t - 1} \B{D}_\z $ \label{step:GradV}
\ENDFOR
\end{algorithmic}
\end{algorithm}

\subsection{Regularizer Induced by DropBlock}
\label{induced-reg}

We first show that the DropBlock Algorithm \ref{alg:sgd} can be interpreted as applying SGD to the objective in \eqref{eq:DropBlockNN}. To that end, recall that the gradient of the expected value is equal to the expected value of the gradient. Thus, the gradient of 
$\| \Y - \frac{1}{\theta} \U (\diag{\w} \otimes \I_{\frac{d}{r}}) \V^\top \X \|_\F^2$ with respect to $\U$ and $\V$ for a random sample of $\w$ provides a stochastic gradient for the objective in \eqref{eq:DropBlockNN}. Steps \ref{step:GradU} and \ref{step:GradV} of Algorithm \ref{alg:sgd} compute such gradients. Therefore, we conclude that the DropBlock Algorithm \ref{alg:sgd} is a SGD method for minimizing \eqref{eq:DropBlockNN}.

The next step is to understand the regularization properties of DropBlock.
The following Lemma\footnote{\noindent Proofs of all our results are given in the Supplementary Material.} shows that the Dropblock
optimization problem is equivalent to a deterministic formulation with a regularization term, which we denote by $\Omega_{\rm DropBlock}$. That is, DropBlock induces explicit regularization.
\begin{lemma} \label{lem:DropBlockDet} The stochastic DropBlock objective \eqref{eq:DropBlockNN} is equivalent to a regularized deterministic objective:
\begin{align}
&\E_{\w} \left \| \Y - \frac{1}{\theta} \U (\diag{\w} \otimes \I_{\frac{d}{r}}) \V^\top \X \right\|_\F^2 \nonumber\\
&\quad = \| \Y - \U  \V^\top \X \|_\F^2 + \Omega_{\rm DropBlock}(\U, \X^\top \V),
\end{align}
where $\Omega_{\rm DropBlock}$ is given by
\begin{equation}
    \Omega_{\rm DropBlock}(\U, \X^\top \V) = \frac{1 - \theta}{\theta} \sum_{i = 1}^k \| \U_i \V^\top_i \X \|_\F^2 \label{DropBlockDet}
\end{equation}
with $\U_i \in \R^{a\times r}$ and $\V_i \in \R^{b \times r}$ denoting the $i^\text{th}$ blocks of $r$ consecutive columns in $\U$ and $\V$ respectively and $k = \frac{d}{r}$ denoting the number of blocks.
\end{lemma}
As expected, when we set $r = 1$, \ie\ when we drop blocks of $1$ neuron independently, $ \Omega_{\rm DropBlock}$ reduces to Dropout regularization in \eqref{eq:DropoutNN-variational}. Therefore, DropBlock regularization generalizes Dropout regularization in \eqref{eq:DropoutNN-variational} by taking the sum over the squared Frobenius norms of rank-$r$ submatrices. But what is the effect of this modification? Specifically, can we characterize the regularization properties of $\Omega_{\rm DropBlock}$, and how it controls the capacity of the network? 

\subsection{Capacity Control Property of DropBlock}
\label{capacity-control}
In this subsection we first study whether DropBlock is capable of constraining the capacity of the network alone. That is, if the network were allowed to be made arbitrarily large, is DropBlock regularization sufficient to constrain the capacity of the network?  

It is clear from the definition of $\Omega_{\rm DropBlock}$ that for any non-zero $(\U,\V)$ the regularizer will be strictly positive. However, it is not clear if the regularizer increases with $d$. The following Lemma shows that when the Dropout probability, $1-\theta$, is constant with respect to $d$, DropBlock alone cannot constrain the capacity of the network, because for any output $\A$ one can find a factorization into $\U \V^\top{\X}$ that makes $\Omega_{\rm DropBlock}$ arbitrarily small (approaching 0 in the limit) by making the width $d$ of the final layer large enough. 
%
\begin{lemma} \label{lemma:zero} Given any matrix $\A$, if the number of columns, $d$, in $(\U,\V)$ is allowed to vary, with $\theta$ held constant, then
\begin{align}
\inf_d \inf_{\substack{\U \in \R^{m\times d},\V \in \R^{n\times d} \\ \A=\U \V^\top \X}} \Omega_{\rm DropBlock}(\U, \X^\top \V) = 0.
\end{align}
\end{lemma}
\noindent
Note that this result is also true for regular Dropout (a special case of DropBlock) with a fixed Dropout probability.

In what follows, we show that if the Dropout probability, $1-\theta$, increases with $d$, then DropBlock is capable of constraining the network capacity.  Specifically, let us denote the retain probability for dimension $d$ as:
\begin{align}
    \theta(d) = \frac{\bar \theta r}{\bar \theta r + (1 - \bar \theta) d}, \label{thetad}
\end{align}
where $\bar \theta = \theta(r)$ denotes the value of the DropBlock parameter when there is only one block, and $d = r$. With $\theta = \theta(d)$, Lemma \ref{lem:DropBlockDet} gives us the following deterministic equivalent of the DropBlock objective:
\begin{equation}
f(\U, \V, d) = \| \Y - \U  \V^\top \X \|_\F^2  +  \frac{d}{r} \frac{1 - \bar\theta}{\bar\theta} \sum_{i = 1}^k \| \U_i \V^\top_i \X \|_\F^2. \label{fuv}
\end{equation}

In order to study the minimizers of $f(\U, \V, d)$, note that at any minimizer $(\U^*, \V^*, d^*)$ we would have the following: 
\begin{equation}
\begin{split}
f(& \U^*,\V^*,d^*) = \|\Y-\U^*\V^{*\top}\X \|_\F^2 + \\
& \quad \inf_d \inf_{\substack{\U \in \R^{a\times d},\V \in \R^{b\times d} \\ \U \V^\top\X =\U^*\V^{*\top}\X}} \frac{d}{r}\frac{1-\bar\theta}{\bar\theta} \sum_{i=1}^k \|\U_i\V_i^\top \X\|_\F^2
\end{split}
\end{equation}
where the last term denotes the fact that given the global minimizer of the matrix product $(\U^* \V^{*\top} \X)$ the regularization induced by DropBlock will induce factors $(\U^*,\V^*)$ which minimize the induced regularization term $\Omega_{\rm DropBlock}$.

This motivates a study of the regularization induced by DropBlock in the product-space, which we denote as $\Lambda(\A)$: 
%
\begin{equation}
\Lambda(\A) = \frac{1 - \bar \theta}{\bar \theta} \inf_d \inf_{\substack{\U \in \R^{a\times d},\V \in \R^{b\times d} \\ \A=\U \V^\top \X}} \frac{d}{r} \sum_{i = 1}^k \|\U_i \V^\top_i \X \|^2_\F. 
\label{dropblockreg} 
\end{equation}
By the definition of $\Lambda(\A)$ in \eqref{dropblockreg}, one can define a function
\begin{equation}
\bar F(\A) = \|\Y-\A \|_F^2 + \Lambda(\A) \label{lowerbd1}
\end{equation}
that globally lower bounds $f(\U,\V, d)$, i.e., 
\begin{equation}
\bar F(\A) \leq f(\U,\V, d), \forall(\U,\V,\A)\! \st \! \U\V^\top \X = \A 
\end{equation}
with equality for $(\U,\V, d)$ that achieve the infimum in \eqref{dropblockreg}.  As a result, $\bar F(\A)$ provides a useful analysis tool to study the properties of solutions to the problem of interest $f(\U,\V, d)$ as it provides a lower bound to our problem of interest in the output space (i.e., $\U\V^\top \X$).

While it is simple to see that $\bar F(\A)$ is a lower bound of our problem of interest, it is not clear whether $\bar F(\A)$ is a useful lower bound or whether the minimizers to $\bar F$ can characterize minimizers of $f$. In the following analysis, we will prove that the answer to both questions is positive. That is, we will show that $\bar F(\A)$ is a tight convex lower bound of $f$, generalizing existing results in the literature \cite{Cavazza:AISTATS18, Mianjy:ICML18}, and that minimizers of $\bar F(\A)$ can be computed in closed form.

\subsection{DropBlock Induces Balanced Weights}
\label{balanced-weights}
In order to characterize the minimizers of $f(\U, \V, d)$, we first need to define the notion of \emph{balanced factors}.
\begin{definition}
A matrix pair $(\U, \V)$ is called \textbf{balanced} if the norms of the products of the corresponding blocks of $\U$ and $\V$ are equal, i.e., $\|\U_1 \V_1^\top \X\|_\F = \|\U_2 \V_2^\top \X\|_\F = \ldots = \|\U_k \V_k^\top \X\|_\F$, where $\U_i$ and $\V_i$ denote the $i^\text{th}$ blocks of $r$ consecutive columns in $\U$ and $\V$ respectively.
\end{definition}

The following result shows that all minimizers of $f(\U,\V, d)$ are balanced. 
\begin{theorem} \label{th:eq}
If $(\U^*, \V^*, d^*)$ is a minimizer of \eqref{fuv}, then $(\U^*, \V^*)$ is balanced. 
\end{theorem}
Theorem \ref{th:eq} provides a characterisation of the minimizers of the DropBlock objective \eqref{fuv}, saying that all the summands in the regulariser are equal at optimality. With this result, we will be able to link the minimizers of $f$ and $\bar F$, and hence find the regularization induced by DropBlock. 

We now note some connections to recent literature. Our result generalizes the balancing result obtained in \cite{Mianjy:ICML18}, which corresponds to the particular case $k=1$. However, our proof technique is radically different. The proof in \cite{Mianjy:ICML18} exploits the rank-$1$ structure to show the existence of an orthonormal matrix $\Q$ such that a given factorization $(\U, \V, d)$  can be transformed to a balanced transformation $(\U \Q, \V \Q, d)$. In contrast, the intuition behind our proof is that when $(\U, \V)$ are not balanced, we can add additional, duplicate blocks of neurons in a particular way to make the block-product-norms $\|\U_i \V_i^\top \X\|_\F$ more balanced, reducing the objective.

Having shown this necessary condition for solutions to $f(\U,\V,d)$, we now use this result to show that $\bar F$ is a tight lower bound of \eqref{fuv}.
\begin{theorem} \label{th:tight}
If $(\U^*, \V^*, d^*)$  is a global minimizer of the factorized problem $f$, then $\A^* = \U^{*} \V^{*\top} \X$ is a global minimizer of the  lower bound $\bar F$. Furthermore, the lower bound is tight, i.e. we have $f(\U^*, \V^*,d^*) = \bar F(\A^*)$. 
\end{theorem}
Theorem \ref{th:tight} provides a link between the hard non-convex problem of interest, $f$, and the lower bound, $\bar F$, and gives us a guarantee that we can verify solutions to $f$ by showing they are solutions to $\bar F$. Hence, we now focus our attention on characterizing solutions of $\bar F(\A)$.

\subsection{DropBlock Induces $k$-support Norm Regularization}
\label{k-support}

Based on the above discussion, we now analyze the global minimizers of $\bar F(\A)$.  Unfortunately, it is not clear yet whether $\Lambda(\A)$ is convex w.r.t. $\A$, which complicates the analysis of the global minimizers of $\bar F(\A)$. Therefore, we will consider instead the lower convex envelope of $\Lambda(\A)$, $\Lambda^{**}(\A)$, and show that it gives a tight bound to the problem in \eqref{fuv}. Furthermore, we will show that $\Lambda(\A)$ (and by extension $\bar F(\A)$) is indeed convex by showing that $\Lambda(\A) = \Lambda^{**}(\A), \ \forall \A$.

First,  recall that the \emph{lower convex envelope} \cite{Rockafellar:2015} of a function $h(x)$ is the largest convex function $g(x)$ such that $\forall x \ g(x) \leq h(x)$, and is given by the Fenchel double dual (i.e., the Fenchel dual of the Fenchel dual). For $\Lambda(\A)$, the following result provides the lower convex envelope. Note that in this sub-section, we will assume that $\X$ has full column rank. This is typically a minor assumption since if $\X$ is not full rank adding a very small amount of noise will make $\X$ full rank. 
\begin{theorem} \label{th:convexlb} When $\X$ has full column rank in \eqref{dropblockreg}, the lower convex envelope of the DropBlock regularizer $\Lambda(\A)$ in \eqref{dropblockreg} is given by
\begin{equation}
\Lambda^{**}(\A) = \frac{1 - \bar \theta}{\bar \theta} \left( \sum_{i = 1}^{\rho^* - 1} a_i^2 + \frac{(\sum_{i = \rho^*}^d a_i)^2}{r - \rho^* + 1} \right), \label{convexlb}
\end{equation}
where $\rho^*$ is the integer in $\{1, 2, \ldots, r\}$ that maximizes \eqref{convexlb}, and $a_1 \geq a_2 \ldots \geq a_d$ are the singular values of $\A$.
\end{theorem}
Note that the quantity $\rho^*$ mentioned in (\ref{convexlb}) is purely a property of the matrix $\A$, the hidden dimension $d$ and the block size $r$, and is determined completely in time $d \log d$, given an SVD of $\A$. 

We again note some connections to recent literature. The form of the solution (\ref{convexlb}) is particularly interesting because it is a matrix norm that has recently been discovered in the sparse prediction literature by \cite{Argyriou:NIPS12}, where it is called the $k$-Support Norm and provides the tightest convex relaxation of sparsity combined with an $\ell_2$ penalty. When applied to the singular values of a matrix (as is the case here), it is called the Spectral $k$-Support Norm, as studied recently in~\cite{Mcdonald:NIPS14}.

\myparagraph{Properties of the $k$-Support Norm}
\label{ksupport}
We are often interested in obtaining sparse or low-rank solutions to problems, as they have been shown to generalize well and are useful in discarding irrelevant features. Specifically, if we are learning a vector $w$, we can get sparse solutions by constraining the $\ell_0$ \emph{norm} of $w$, that is the number of non-zero entries in $w$. However, $\| \cdot \|_0$ is not a convex function (and hence not a norm), and it is hard to solve an optimization problem with the constraint set $S_0 = \{w : \|w\|_0 \leq k\}$.
Hence, typically we relax the regularizer to be the $\ell_1$ norm, which has nicer properties. Constraining the $\ell_1$ norm does not yield a convex relaxation of $S_0$, in the sense that $\|w\|_0$ might be small while $\|w\|_1$ is large. However, additionally constraining the $\ell_2$ norm fixes this problem, as the convex hull of the set $S_{0, 2} = \{w : \|w\|_0 \leq k, \|w\|_2 \leq 1\}$ is a subset of  $S_{1, 2} = \{w : \|w\|_1 \leq \sqrt{k}, \|w\|_2 \leq 1\}$, i.e. ${\rm conv}(S_{0, 2}) \subseteq S_{1, 2}$. This motivates the use of the elastic-net regularizer in literature. Recently, researchers have looked at whether $S_{1, 2}$ is the \emph{tightest} convex relaxation of $S_{0, 1}$, and found that it is not. Specifically, \cite{Argyriou:NIPS12} show that this tightest convex envelope can be obtained in closed form as a norm, which they call the $k$-Support norm of $w$.

The $k$-Support Norm is essentially a trade-off between an $\ell_2$ penalty on the largest components, and an $\ell_1$ penalty on the remaining smaller components. In our case, when $\rho^* \!= \!1$ in \eqref{convexlb}, $\Lambda^{**}(\A)$  reduces to $\frac{c_0}{r} (\sum_{i = 1}^d a_i)^2 \!=\! \frac{c_0}{r} \| \A \|_*^2$, which is (a scaling of) the nuclear norm (squared) of $\A$. 
On the other hand, when the block size $r$ is larger, $\rho^*$ will take higher values, implying the regularizer $\Lambda^{**}(\A)$ will move closer to $c_0 \sum_{i = 1}^d a_i^2 = c_0 \|\A\|_\F^2$, which is (a scaling of) the squared Frobenius norm of $\A$. Therefore, the DropBlock regularizer acts as an interpolation between (squared) nuclear norm regularization when the block size is small to (squared) Frobenius norm regularization when the block size becomes very large. Further, \cite{Argyriou:NIPS12, Mcdonald:NIPS14} observe that regularization using the $k-$Support norm achieves better performance than other forms of regularization on some real-world datasets and this might be a step towards theoretically explaining the superior performance of DropBlock compared to Dropout, as was observed experimentally~in~\cite{Ghiasi:NIPS18}.

\myparagraph{Closed Form Solutions for DropBlock}
Continuing our analysis, with the convex envelope of $\Lambda(\A)$, we can construct a convex lower bound of the DropBlock objective $f(\U, \V, d)$, as follows:
\begin{equation}
\F(\A) = \|\Y - \A\|_\F^2 + \Lambda^{**}(\A) \label{lowerbd}
\end{equation}
with the relationship $\F(\A) \leq \bar F(\A) \leq f(\U,\V)$ for all $(\U,\V,\A)$ such that $\U \V^\top = \A$. 
As shown earlier in Theorem \ref{th:tight} the lower bound $\bar F(\cdot)$ takes the same minimum value as $f(\cdot, \cdot, \cdot)$. By properties of the lower convex envelope, we know that the function $\bar F(\cdot)$ takes the same minimum value as its convex lower bound $\F(\cdot)$. Following this reasoning, we now complete the analysis by deriving a closed form solution for the global minimum of $\F(\A)$.%
\begin{theorem} \label{th:globmin} When $\X$ has full column rank in \eqref{dropblockreg}, the global minimizer of $\F(\A)$ is given by $\A_{\rho, \lambda} = \U_\Y \diag{\a_{\rho, \lambda}} \V_\Y^\top$, where $\Y = \U_\Y \diag{\m} \V_\Y^\top$ is an SVD of $\Y$, and $\a_{\rho, \lambda}$ is given by
%
\begin{equation}
\a_{\rho, \lambda}=
\begin{cases}
\left( \frac{m_1}{\beta + 1}, \frac{m_2}{\beta + 1}, \ldots, \frac{m_\lambda}{\beta + 1}, 0, \ldots, 0 \right) 
& \text{if } \lambda \leq \rho - 1\\
\left(
\begin{array}{ll}
\frac{m_1}{\beta + 1}, \frac{m_2}{\beta + 1}, \ldots, \frac{m_{\rho - 1}}{\beta + 1}, \\
m_{\rho} - \frac{\beta}{c} S, m_{\rho + 1} - \frac{\beta}{c} S, \ldots, \\
m_{\lambda} - \frac{\beta}{c} S, 0, \ldots, 0
\end{array}
\right)
& \text{if } \lambda \geq \rho. 
\end{cases} \nonumber
\end{equation}

The constants are $\beta = \frac{1 - \bar \theta}{\bar \theta}$, $S = \sum_{i = \rho}^\lambda m_i$, $c = r + \beta \lambda + (\beta + 1) (1 - \rho)$, while  $\rho \in \{1, 2, \ldots, r - 1\}$ and $\lambda \in \{1, 2, \ldots, d\}$ are chosen such that they minimize $\F(\A_{\rho, \lambda})$.
\end{theorem}

Note that the constants mentioned in Theorem \ref{th:globmin} depend purely on the matrix $\Y$, and can be computed in time $O(d^2)$ given the singular values $m_i$. Finally, the following Corollaries obtained from Theorem \ref{th:eq} complete the picture by showing that the solution computed in Theorem \ref{th:globmin} recovers the value of the global minimizer of the DropBlock objective $f(\cdot, \cdot, \cdot)$, and that $\Lambda$ is convex:
\begin{corollary} \label{cor:allequal}
If $\A^*$ is a global minimizer of the lower convex envelope $\F$, and $(\U^*, \V^*, d^*)$ is a global minimizer of the non-convex objective $f$, then we have $\F(\A^*) = \bar F(\A^*) = f(\U^*, \V^*, d^*)$ with $\A^* = \U (\V^*)^\top \X$. 
\end{corollary}
\begin{corollary} \label{cor:lambdaconvex}
$\Lambda(\A)$ is convex and equal to its lower convex envelope $\Lambda^{**}(\A)$ (i.e., $\Lambda(\A) = \Lambda^{**}(\A), \forall \A)$.
\end{corollary}
Note that the Corollary \ref{cor:allequal} also trivially applies if $\A^*$ is a global minimizer of $\bar F$ since $F$ and $\bar F$ share the same set of global minimizers. Having understood the properties of one particular generalization of dropout for a single hidden-layer linear network, we will now show how our methods can be generalised to other Dropout variants applied to the last layer of an overparameterized neural network.

\section{Generalized Dropout Framework}
\label{generalised}

In practice, commonly used neural network architectures typically have a fully-connected linear layer as the final layer in the network, and it is also common to use Dropout-style regularization only on this final fully-connected layer.  This leads us to consider the effect of training deep NNs with Dropout-style regularization applied to a final linear layer.  Specifically, we will consider a NN training problem with squared-loss of the form
%
\begin{equation}\label{dropoutNNgeneral}
\min_{\U,\Gamma} \E_{\z} \left \| \Y - \U \diag{\boldsymbol{\mu}}^{-1} \diag{\z} g_\Gamma(\X) \right\|_\F^2,
\end{equation}
where we follow the notation in the introduction, and in addition we let $g_\Gamma$ denote the output of the second to last layer of a NN with weight parameters $\Gamma$ (i.e., the $j^\text{th}$ column of $g_\Gamma$ is the output of second to last layer of the network given input $\x_j$), $\U \in \R^{a \times d}$ be the weight matrix for the final linear layer, with $d$ being the size of the output of the second to last layer, and $\bmu \in \R^d \backslash \0$ be a vector of the means of the Dropout variables, $\mu_i = \E[z_i]$ (note that in expectation, the output of the $i^\text{th}$ hidden unit of $g_\Gamma$ is scaled by $\E[z_i]$, so to counter this effect, we rescale the output by $\E[z_i]^{-1}$).

We will assume that the Dropout variables $\z$ are stochastically sampled at each iteration of the algorithm from an arbitrary probability distribution $\mathcal{S}$ with covariance matrix $ \C = {\rm Cov}(\z, \z)$ and mean $\bmu$. Assuming that each entry of $\bmu$ is non-zero, we define the Characteristic Matrix $\bar \C$ from entries of the mean and covariance of $\z$, $\mu_i$ and $c_{i,j}$, as
\begin{equation} 
\bar c_{i, j} = \frac{c_{i, j}}{\mu_i \mu_j} \quad\text{or}\quad \bar\C = \diag{\bmu}^{-1}\C\diag{\bmu}^{-1}.
\label{cbar}
\end{equation}
Recall that one iteration of a typical Dropout algorithm can be interpreted as performing one iteration of stochastic gradient descent on \eqref{dropoutNNgeneral}, where the gradient of \eqref{dropoutNNgeneral} is approximated by a single stochastic sample of the Dropout variables, $\z$. In this setting, we can obtain the deterministic form of \eqref{dropoutNNgeneral}, which is a generalisation of Lemma \ref{lem:DropBlockDet}:
\begin{lemma} \label{th:gendet} The Generalized Dropout objective (\ref{dropoutNNgeneral}) is equivalent to a regularized deterministic objective:
\begin{align}
&\E_{\z} \left \| \Y - \U \diag{\boldsymbol{\mu}}^{-1} \diag{\z} g_\Gamma(\X) \right\|_\F^2 \nonumber \\
&\quad = \left \| \Y - \U g_\Gamma(\X) \right\|_\F^2 + \Omega_{\C, \bmu}(\U, g_\Gamma(\X)^\top) ,\label{gendet}
\end{align}
where the ``generalized Dropout'' regularizer is defined as
\begin{align}
    \Omega_{\C, \bmu} (\U,\V) &= \sum_{i, j = 1}^d c_{i, j} \frac{(\u_i^\top \u_j) (\v_i^\top \v_j)}{\mu_i \mu_j} \nonumber \\
    &= \langle \bar \C, \U^\top \U \odot \V^\top \V \rangle ,\label{genreg}
\end{align}
with $\u_i$ and $\v_i$ denoting the $i^\text{th}$ columns of $\U$ and $\V$, resp.
\end{lemma}
Note we have defined $\Omega_{\C,\bmu}$ for general matrices $(\U,\V)$ for notational simplicity, but typically we will have $\V = g_\Gamma(\X)^\top$). 
Notice also that $\bar \C$ completely determines the regularization properties of any dropout scheme. For example, in classical Dropout, the entries of $\z$ are i.i.d. Bernoulli variables with mean $\theta$, hence $\C$ is diagonal with diagonal entries $c_{i,i} = \theta(1-\theta)$ and $\mu_i = \theta$, hence $\bar \C$ is diagonal with diagonal entries $\bar c_{i,i} = \frac{1 - \theta}{\theta}$. For DropBlock with block-size $r$, we have $\bar \C = \tfrac{1-\theta}{\theta} \text{BlkDiag} (\1_r \1_r^\top, \ldots, \1_r \1_r^\top)$, where $\text{BlkDiag}(\cdot)$ denotes forming a block diagonal matrix with the function arguments along the diagonal and $\1_r$ denotes an $r$-dimensional vector of all ones.
In the case of Dropout, we recover an immediate simple corollary for the regularization induced by Dropout in the final layer of non-linear networks:
\begin{corollary} \label{cor:dropoutextend}
For regular Dropout applied to objective \eqref{dropoutNNgeneral} the following equivalence holds:
\vspace{-2mm}
\begin{align} \label{dropoutDNNreg}
&\E_{\z} \left \| \Y - \U \diag{\boldsymbol{\mu}}^{-1} \diag{\z} g_\Gamma(\X) \right\|_\F^2 \nonumber \\
&\quad = \left \| \Y - \U g_\Gamma(\X) \right\|_\F^2 + \sum_{i=1}^d \|\u_i\|_2^2 \|g_\Gamma^i(\X)\|_2^2,
\vspace{-2.5mm}
\end{align}
where $g_\Gamma^i(\X) \in \R^N$ denotes the output of the $i^\text{th}$ neuron of $g_\Gamma$ (i.e., the $i^\text{th}$ row of $g_\Gamma(\X)$). 
\end{corollary}

Given this result, a simple interpretation of Dropout in the final layer of the network is that it adds a form of weight-decay both to the weight parameters in the final layer, $\U$, and the output of $g_\Gamma$.  Additionally, from this result it is relatively simple to show the following characterization of the regularization induced by Dropout applied to the final layer of a network. Note that the following result (Proposition 
\ref{thm:nucnorm_limit}) can be shown using similar arguments to those used in \cite{Cavazza:AISTATS18} along with a sufficient capacity assumption.

\begin{proposition}
\label{thm:nucnorm_limit}
If the network architecture, $g_\Gamma$, has sufficient capacity to span $\R^{d \times N}$ (i.e., for all $\Q \in \R^{d \times N}$ there is a set of network weights $\bar \Gamma$ such that $g_{\bar \Gamma}(\X) = \Q$) and $d \geq \min\{a,N\}$, then the global optimum of \eqref{dropoutNNgeneral} with $\z \overset{\text{i.i.d}}{\sim}$ Bernoulli$(\theta)$ is given by:
\vspace{-2mm}
\begin{align}
&\min_{\U,\Gamma} \E_{\z} \left \| \Y - \U \diag{\boldsymbol{\mu}}^{-1} \diag{\z} g_\Gamma(\X) \right\|_\F^2 \nonumber \\
&\quad = \min_{\A} \|\Y - \A\|_F^2 + \tfrac{1-\theta}{\theta} \|\A\|_*^2
\vspace{-2mm}
\end{align}
where $\|\A\|_*$ denotes the nuclear norm of $\A$.
\end{proposition}

The above result has interesting implications in the sense that it implies that even in the limit where the $g_\Gamma$ network has infinite capacity and can represent an arbitrary output perfectly, applying Dropout to the final layer still induces capacity constraints on the output of the overall network in the form of (squared) nuclear norm regularization, where the strength of the regularization depends on the Dropout rate $(1-\theta)$. A result similar to Proposition \ref{thm:nucnorm_limit} can be obtained for DropBlock applied to the last layer of a network with sufficient capacity (the sampling strategy for $\z$ changes and the regularizer changes from nuclear norm squared to $k$-support norm squared). Having analyzed Dropout and its variants, we now consider an alternative but closely related approach, DropConnect in the next section.

\vspace{-1mm}
\section{DropConnect Analysis}
\vspace{-1mm}
\label{dropconnect}
DropConnect, proposed in \cite{Wan:ICML13}, is very similar to Dropout, but instead of setting the outputs of hidden neurons to zero, DropConnect instead sets \emph{elements} of the connection weights to zero independently with probability $1 - \theta$. Hence, the DropConnect algorithm samples a random matrix $\Z \in \R^{b \times d}$, with each $z_{i, j}$ drawn independently from the Bernoulli distribution with parameter $\theta$. 
For Dropconnect applied to the second-last layer weights $\V$ of a deep network parameterized as $\U \V^\top g_\Gamma(\X)$, the optimization problem then becomes the following:
\vspace{-1.5mm}
\begin{equation}\label{dropconnectDeepNN}
\min_{\U,\V,\Gamma} \E_{\Z} \left \| \Y - \frac{1}{\theta} \U(\Z \odot \V)^\top g_\Gamma(\X) \right\|_\F^2
\vspace{-1mm}
\end{equation}
Note that we apply DropConnect to the second-last layer $\V$ instead of $\U$ in order to match the original authors proposal~\cite{Wan:ICML13}. We show that DropConnect induces the same regularization as Dropout. Specifically, the regularizer induced in (\ref{dropconnectDeepNN}) is the same as applying vanilla Dropout on the last layer:
\begin{theorem} \label{th:dropconnectdnn}
For Dropconnect applied to the second-to-last layer weights $\V$ of a deep network parameterized as $\U \V^\top g_\Gamma(\X)$, the following equivalence holds:%
\vspace{-2mm}
\begin{multline}%
\E_{\Z} \left \| \Y - \frac{1}{\theta} \U(\Z \odot \V)^\top g_\Gamma(\X) \right\|_\F^2 \\
= \left \| \Y - \U \V^\top g_\Gamma(\X) \right\|_\F^2 + \frac{1 - \theta}{\theta}\sum_{i = 1}^{d} \|\u_i\|_2^2 \|g_\Gamma(\X)^\top \v_i\|_2^2 \nonumber
\end{multline}%
where $g_\Gamma^i(\X) \in \R^N$ denotes the output of the $i^\text{th}$ neuron of $g_\Gamma$ (i.e., the $i^\text{th}$ row of $g_\Gamma(\X)$). 
\end{theorem}

Taking $g_\Gamma(\X) = \X$ in Theorem \ref{th:dropconnectdnn} then gives us the following result for a single layer linear network.
\begin{corollary} \label{equivalence}
For single layer linear networks, the stochastic DropConnect objective \eqref{dropconnectDeepNN} is equivalent to the vanilla Dropout deterministic objective:
	\begin{align}
	&\mathbb{E}_{\mathbf{Z}} \left \| \Y - \frac{1}{\theta} \U(\Z \odot \V)^\top \X \right\|_2^2 \nonumber \\
	&= \| \Y - \U  \V^\top \X \|_2^2 + \frac{1 - \theta}{\theta}\sum_{i = 1}^{d} \|\u_i\|_2^2 \|\X^\top \v_i\|_2^2 \label{dropconnectNNdet}
	\end{align}
\end{corollary}
Note that by an identical line of arguments as made in \cite{Cavazza:AISTATS18,Mianjy:ICML18} the above result also implies that DropConnect induces low-rank solutions in linear networks. 

\section{Experiments}
\label{experiments}
\begin{figure}
\centering
\includegraphics[width=0.3\textwidth]{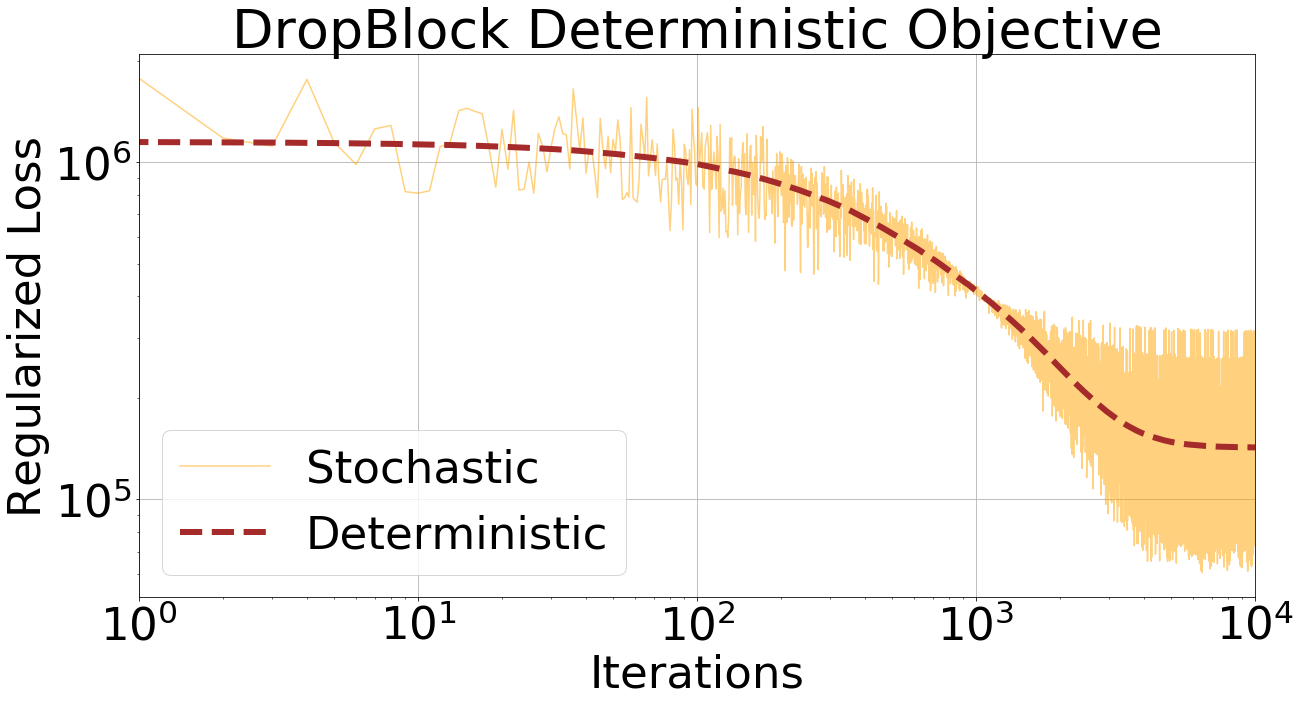}
\includegraphics[width=0.3\textwidth]{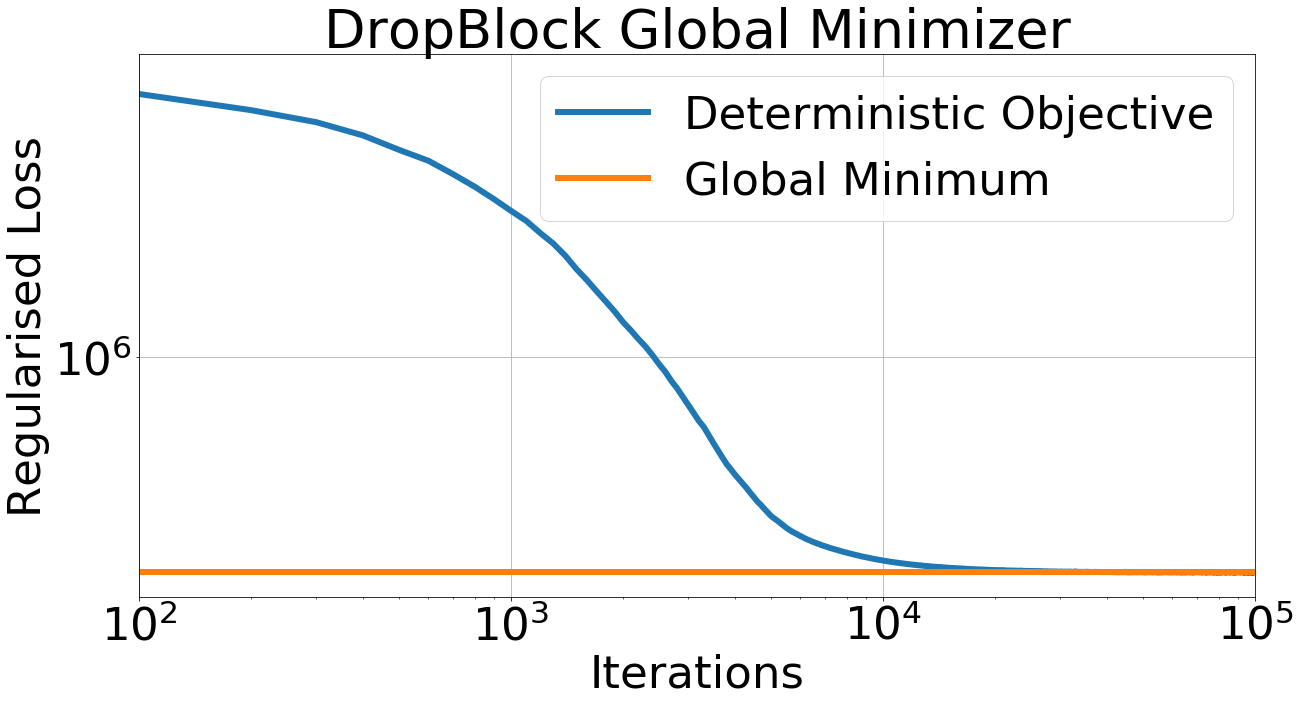}
\centering
\caption{\label{fig:dropblock} Top: Stochastic DropBlock training with SGD is equivalent to the deterministic objective \eqref{DropBlockDet}. Bottom: DropBlock converges to the global minimum computed in Theorem \ref{th:globmin}.}
\end{figure}

\begin{figure}
\centering
\includegraphics[width=0.3\textwidth]{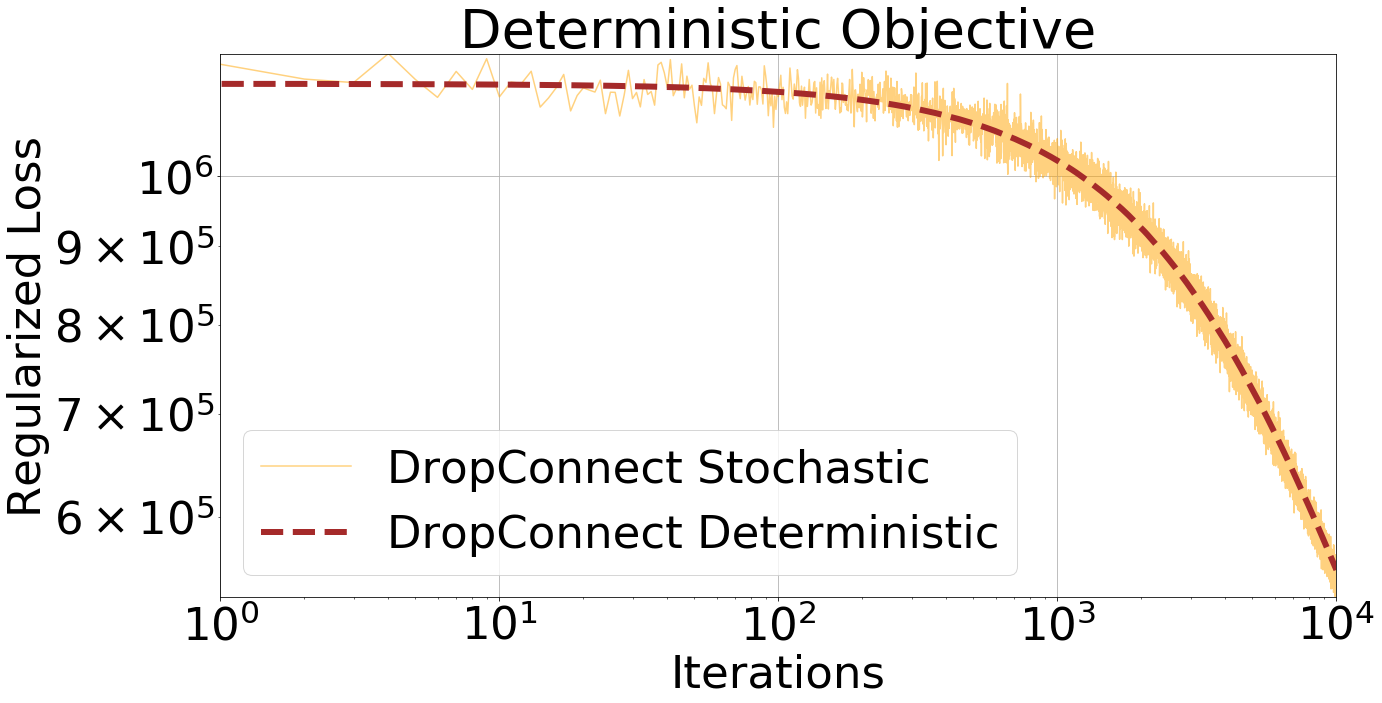}
\includegraphics[width=0.3\textwidth]{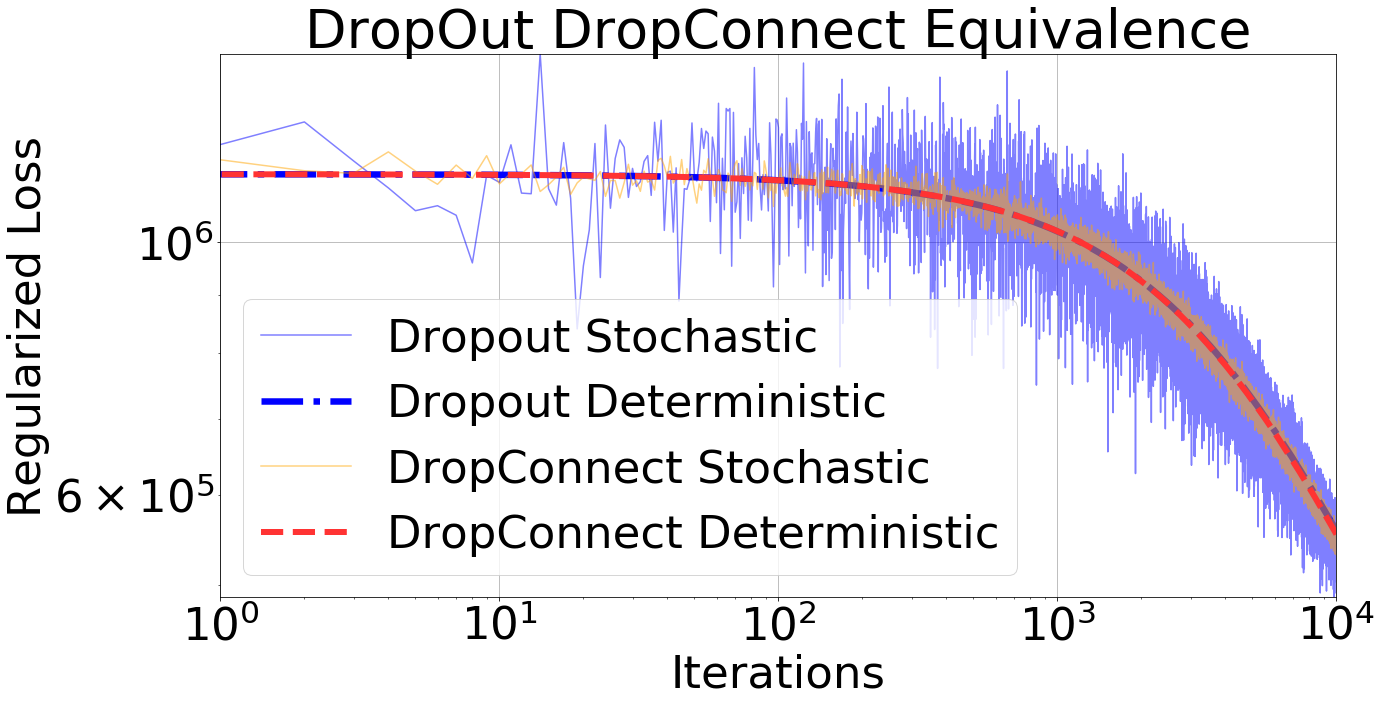}
\caption{\label{fig:dropconnect} Comparing DropConnect to DropOut. Top: Stochastic DropConnect training with SGD is equivalent to the deterministic Objective \eqref{dropconnectNNdet}. Bottom: DropConnect training is equivalent to Dropout training for the squared loss.}
\end{figure}

In this section, we conduct experiments in training single hidden layer linear networks as well as multilayer nonlinear networks to validate the theory developed so far. 

\subsection{Shallow Network Experiments}
We first create a simple synthetic dataset $\mathcal{D}_{\rm syn}$ by taking $1000$ i.i.d samples of $\x$ from a $100$-dimensional standard normal distribution. Then, $\y \in \R^{80}$ is generated as $\y = \M \x$, where $\M = \U_{\rm true} \V_{\rm true}^\top$. To ensure a reliable comparison, all the experiments 
start with the same choice of $\U_0 = \U_{\rm init} \in \R^{80 \times 50}$ and $\V_0 = \V_{\rm init} \in \R^{100 \times 50}$. The entries of all the matrices $\U_{\rm true}, \V_{\rm true}, \U_{\rm init}, \V_{\rm init}$ are sampled elementwise from $\mathcal{N}(0, 1)$.
\vspace{-0.1cm}

\myparagraph{Verifying Deterministic Formulations} We first verify the correctness of the deterministic formulations for various dropout schemes analyzed in this paper, i.e. \eqref{DropBlockDet} and \eqref{dropconnectNNdet}, in the top panels of Figure \ref{fig:dropblock} and Figure \ref{fig:dropconnect}. In Figure \ref{fig:dropblock}, the curve labelled \emph{DropBlock Stochastic} is the training objective plot, \ie it plots $\left \| \Y - \frac{1}{\theta} \U_t (\diag{\w_t} \otimes \I_{\frac{d}{r}}) \V_t^\top \X \right\|_\F^2$ 
as the training progresses via Algorithm \ref{alg:sgd}.
For generating the curve labeled \emph{DropBlock Deterministic}, we take the current iterate, \ie \ $\U_t, \V_t$, and plot the Deterministic  DropBlock objective obtained in Lemma \ref{lem:DropBlockDet} at every iteration. The deterministic equivalent of the DropConnect objective is similarly verified in Figure \ref{fig:dropconnect}. Both the figures show plots for $\theta = 0.5$, and the plots for more values of $\theta$ are deferred to the Appendix. It can be seen that the expected value of DropConnect and DropBlock over iterations matches the values derived in our results. Additionally, the bottom panel of Figure \ref{fig:dropconnect} shows that Dropout and DropConnect have the same expected value of the objective at each iteration.

\myparagraph{Verifying Convergence to the Global Minimum}%
We next verify the convergence of DropBlock to the theoretical global minimum computed in Theorem \ref{th:globmin}. The bottom panel of Figure \ref{fig:dropblock} plots the deterministic DropBlock objective as the training progresses, showing convergence to the computed theoretical global minimum. It can be seen that the training converges to the DropBlock Global minimum computed in Theorem \ref{th:globmin}.

\begin{figure}
\centering
\includegraphics[width=0.33\textwidth]{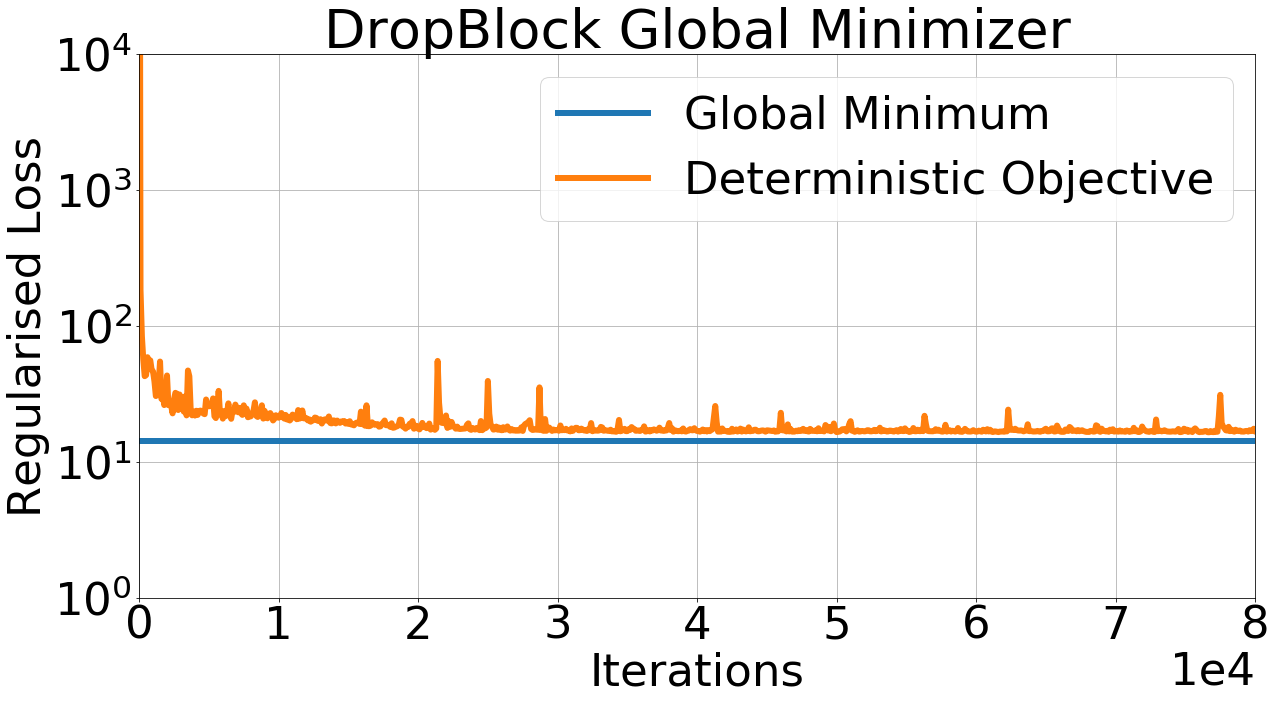}
\includegraphics[width=0.33\textwidth]{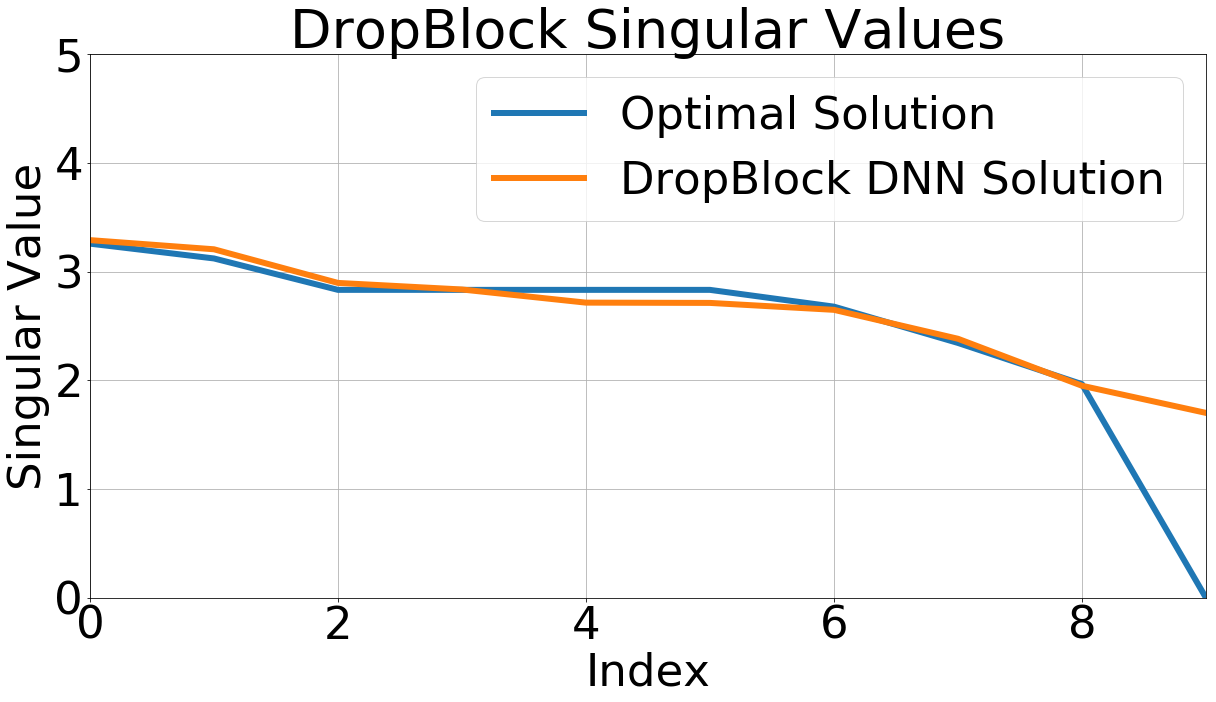}
\includegraphics[width=0.33\textwidth]{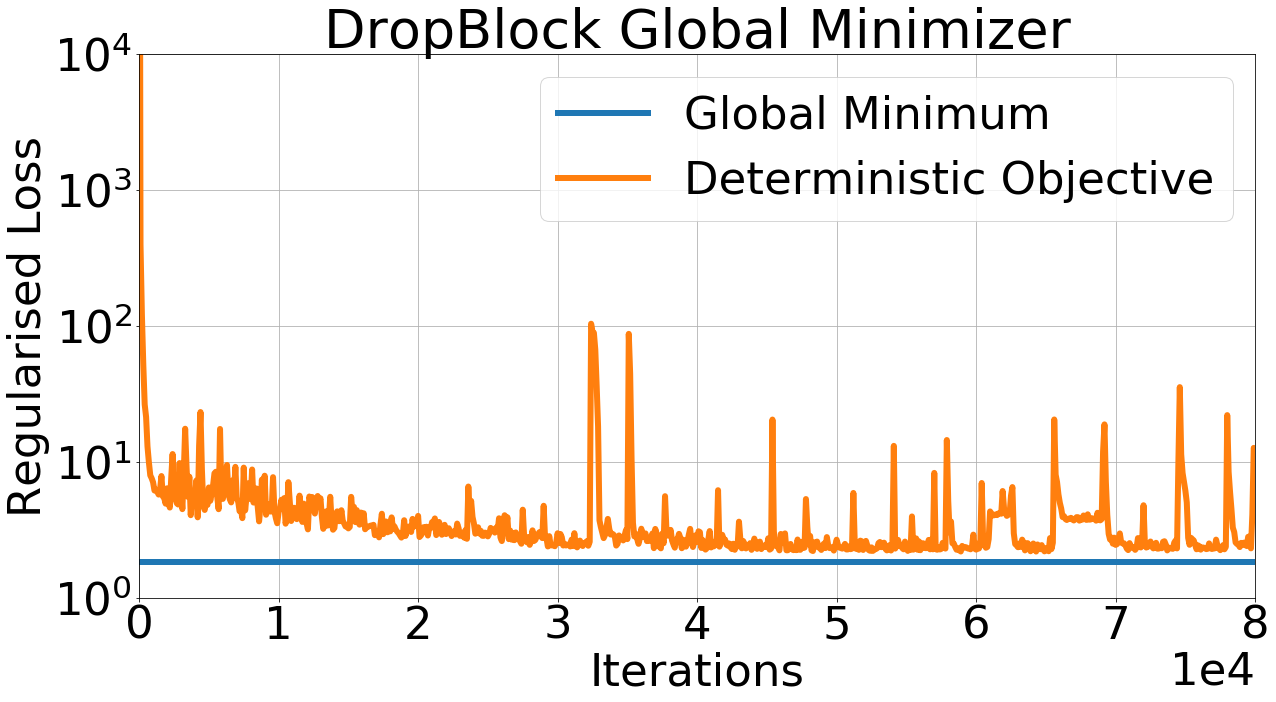}
\includegraphics[width=0.33\textwidth]{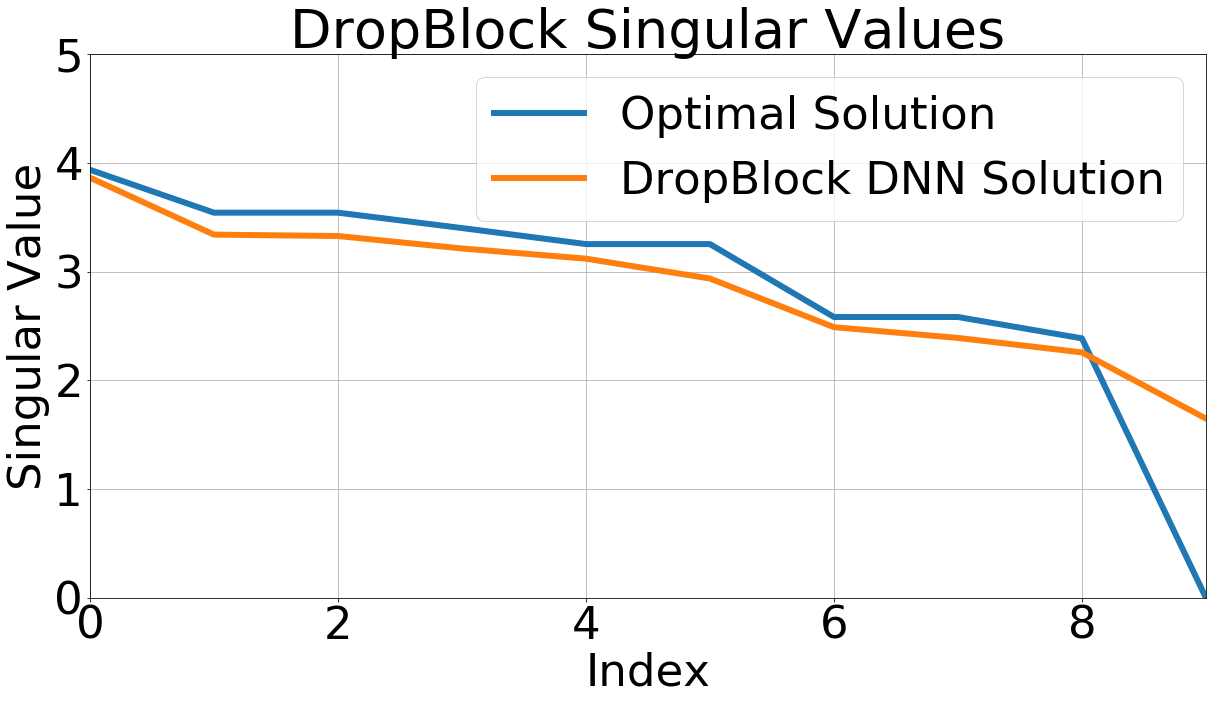}
\caption{
    \label{fig:syn1} Results for Resnet-50 training on MNIST (first and second panels) and CIFAR10 (third and fourth panels) data.  The first and third panels show the deterministic loss during each training iteration as training progresses, and the second and fourth panels show the singular values of the product matrix of the final iterate. 
    }
\end{figure}
\subsection{Deep Network Experiments} 
In order to test our predictions on common network architectures, we modify the standard Resnet-50 architecture by removing the last layer and inserting a fully-connected (FC) layer to reduce the hidden layer dimensionality to $80$ (to make the experiments consistent with the Synthetic Experiments). Hence, the network architecture now is, $\x \rightarrow \texttt{ Resnet-50 Layers} \rightarrow \texttt{FC} \rightarrow \texttt{Dropout}  \rightarrow \texttt{FC} \rightarrow \y$. We then train the entire network on small datasets $\mathcal{D}_{\rm MNIST}, \mathcal{D}_{\rm CIFAR10}$ 
with DropBlock applied to the last layer with a block size of $5$. Figure \ref{fig:syn1} shows that the solution
found by gradient descent is very close to the lower bound predicted by Theorem \ref{th:globmin}: The objective value is plotted on the first and third panels, and the singular values of the final predictions matrix $\U g_\Gamma(\X)$ are plotted in decreasing order on the second and fourth panels.  Note that qualitatively the singular values of the final predictions matrix closely match the theoretical prediction, with the exception of the least significant singular value, which we attribute to the highly non-convex network training problem not converging completely to the true global minimum.

\subsection{Effect of DropBlock approximation}
\label{approxverify}
\label{sec:approximation}
\begin{figure}
\centering
\includegraphics[width=0.40\textwidth]{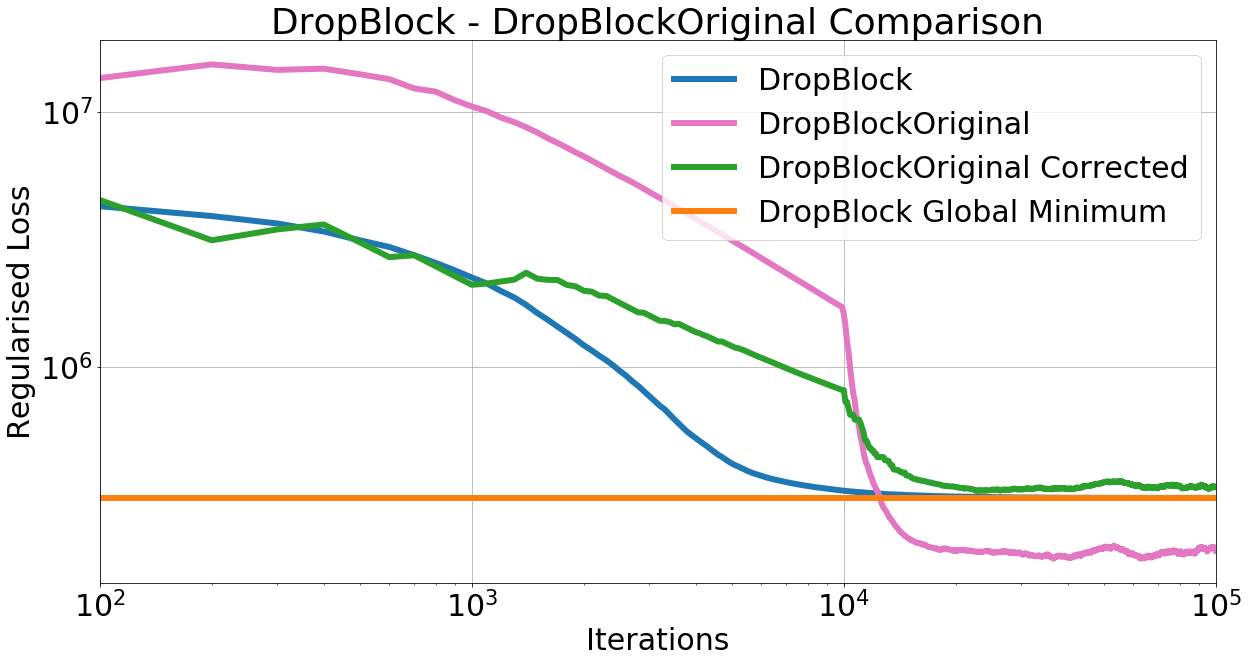}
\vspace{-1.5mm}
\caption{Comparison of DropBlockOriginal and the approximation DropBlock we have made. The training curves correspond to $\theta_{\rm DropBlock} = 0.5$ of Fig \ref{fig:dropblock}. The curves have been smoothed via an exponential moving average.}
 \label{fig:comparison}
\end{figure}

The original DropBlock method \cite{Ghiasi:NIPS18} allows dropping blocks at arbitrary locations, in this paper we made an approximation by constraining the blocks to be non-overlapping, as mentioned in the beginning of Section \ref{DropBlock}. 
This approximation is a minor constraint, and the block retaining probability $\theta$ can be scaled appropriately to recover the original behavior. DropBlockOriginal with the same $\theta$ as DropBlock leads to a higher \emph{effective} dropping rate. This can be corrected by solving for $\theta'_{\rm DBOriginal}$ such that the probability of dropping any neuron in DropBlock with retain probability $\theta_{\rm DropBlock}$ is same as the probability of dropping a neuron in DropBlockOriginal with retain probability $\theta'_{\rm DBOriginal}$. Specifically, referring to the notation in Section \ref{DropBlock}, under the Original DropBlock scheme, the probability of $z_i = 0$ is same as the probability of (none of the $w_j = 1$) over all $j$, where $|i - j| \leq k$. This probability is $(1 - \theta'_{\rm DBOriginal})^{2k - 1}$. Under our approximation, the probability of $z_i = 0$ is $1 - \theta_{\rm DropBlock}$. Equating these quantities, we can solve for $\theta'_{\rm DBOriginal}$ as $\theta'_{\rm DBOriginal} = 1 \!-\! (1 \!-\! \theta_{\rm DropBlock})^{\frac{1}{2k \!-\! 1}}$.
%
As can be seen in Fig \ref{fig:comparison}, DropBlockOriginal with the appropriate correction is approximately the same as DropBlock, as the green, blue, orange curves are very close in log-scale at iteration $10^5$.

\section{Conclusion}
\label{conclusion}
In this work, we have analysed the regularization properties of structured Dropout training of neural networks, and characterized the global optimum obtained for some classes of networks and structured Dropout strategies. We showed that DropBlock induces spectral $k$-Support norm regularization on the weight matrices, providing a potential way of theoretically explaining the empirically observed superior performance of DropBlock as compared to Dropout. We also proved that Dropout training is equivalent to DropConnect training for some network classes. Finally, we showed that our techniques can be extended to other generic Dropout strategies, and to Deep Networks with Dropout-style regularization applied to the last layer of the network, significantly generalizing prior results.
\vspace{-2mm}
\paragraph{Acknowledgements} This work was supported by IARPA contract D17PC00345 and NSF Grants 1618485 and 1934979. 

{\small
\bibliographystyle{ieee_fullname}
\bibliography{biblio/learning,biblio/vidal}

\begin{thebibliography}{10}\itemsep=-1pt

\bibitem{Argyriou:NIPS12}
Andreas Argyriou, Rina Foygel, and Nathan Srebro.
\newblock Sparse prediction with the $ k $-support norm.
\newblock In {\em Advances in Neural Information Processing Systems}, pages
  1457--1465, 2012.

\bibitem{Cavazza:AISTATS18}
J. Cavazza, B.D. Haeffele, C. Lane, P. Morerio, V. Murino, and R. Vidal.
\newblock Dropout as a low-rank regularizer for matrix factorization.
\newblock In {\em International Conference on Artificial Intelligence and
  Statistics}, volume~84, pages 435--444, 2018.

\bibitem{Gal:NIPS17}
Yarin Gal, Jiri Hron, and Alex Kendall.
\newblock Concrete dropout.
\newblock In {\em Advances in Neural Information Processing Systems}, 2017.

\bibitem{Gastaldi:Arxiv17}
Xavier Gastaldi.
\newblock Shake-shake regularization.
\newblock In {\em arXiv preprint arXiv:1705.07485}, 2017.

\bibitem{Ghiasi:NIPS18}
Golnaz Ghiasi, Tsung-Yi Lin, and Quoc~V Le.
\newblock Dropblock: A regularization method for convolutional networks.
\newblock In {\em Advances in Neural Information Processing Systems}, pages
  10750--10760, 2018.

\bibitem{Haeffele:PAMI2019}
Benjamin~David Haeffele and Ren{\'e} Vidal.
\newblock Structured low-rank matrix factorization: Global optimality,
  algorithms, and applications.
\newblock {\em {IEEE} transactions on pattern analysis and machine
  intelligence}, 2019.

\bibitem{Mcdonald:NIPS14}
Andrew~M McDonald, Massimiliano Pontil, and Dimitris Stamos.
\newblock Spectral k-support norm regularization.
\newblock In {\em Advances in Neural Information Processing Systems}, pages
  3644--3652, 2014.

\bibitem{Mianjy:ICML18}
Poorya Mianjy, Raman Arora, and Rene Vidal.
\newblock On the implicit bias of dropout.
\newblock In {\em International Conference on Machine Learning}, 2018.

\bibitem{Morerio:ICCV17}
P. Morerio, J. Cavazza, R. Volpi, R. Vidal, and V. Murino.
\newblock Curriculum dropout.
\newblock In {\em {IEEE} International Conference on Computer Vision}, Oct
  2017.

\bibitem{Rennie:IEEESLT14}
Steven~J Rennie, Vaibhava Goel, and Samuel Thomas.
\newblock Annealed dropout training of deep networks.
\newblock In {\em 2014 IEEE Spoken Language Technology Workshop (SLT)}, 2014.

\bibitem{Rockafellar:2015}
Ralph~Tyrell Rockafellar.
\newblock {\em Convex analysis}.
\newblock Princeton university press, 2015.

\bibitem{Wan:ICML13}
Li Wan, Matthew Zeiler, Sixin Zhang, Yann Le~Cun, and Rob Fergus.
\newblock Regularization of neural networks using dropconnect.
\newblock In {\em International Conference on Machine Learning}, pages
  1058--1066, 2013.

\bibitem{Yamada:Arxiv18}
Yoshihiro Yamada, Masakazu Iwamura, Takuya Akiba, and Koichi Kise.
\newblock Shakedrop regularization for deep residual learning.
\newblock In {\em arXiv preprint arXiv:1802.02375}, 2018.

\bibitem{Zolna:Arxiv17}
Konrad Zolna, Devansh Arpit, Dendi Suhubdy, and Yoshua Bengio.
\newblock Fraternal dropout.
\newblock In {\em arXiv preprint arXiv:1711.00066}, 2017.

\end{thebibliography}
}

\onecolumn 
\newpage
\appendix
\section{Notation and Assumptions} 
\label{notation}

Matrices are denoted by boldface uppercase letters $\Z$, vectors by boldface lowercase letters $\z$ and scalars by lowercase letters $z$. Unless otherwise stated, scalars and their corresponding matrices and vectors are represented by the same character. For example, scalars $z_{i, j}$ make up the vector $\z_j$ and the vectors $\z_j$ make up the columns of the matrix $\Z$. $\I_d$ represents the identity matrix having dimensions $d \times d$. The subscript form $\z_k$ on a matrix $\Z$ denotes the $k^{\text{th}}$ column of $\Z$. Given a matrix $\Z$, $\Z_i$ will denote the $i^{\text{th}}$ submatrix of $\Z$ formed by sampling a set of columns from $\Z$, where the sampling will be clear from context.  A colon in the subscript, $\z_{i:j}$ denotes a vector formed by the elements $z_i, z_{i + 1}, \ldots, z_{j - 1}, z_j$. For NNs, we the weight matrices are notated with $\U, \V$, with the output of the neural network in response to an input $\x$ being $\U \V^\top \x$, and $\Y$ will denote the target outputs $\Y$ of the network given training data, $\X$, as the input. $d$ will be the number of units in the hidden layer hidden layer, $b$ will denote the input-data dimension, and $a$ will be the dimension of the output. Hence, $\U \in \R^{a \times d}$, $\V \in \R^{b \times d}$. The Hadamard product and the Kronecker product are denoted by $\odot$ and $\otimes$, respectively. Given a matrix $\M$, the Frobenius norm of the matrix is denoted by $\| \M \|_\F$, and the inner product between two matrices $\B{M}, \B{N}$ is denoted by $\langle \B{M}, \B{N} \rangle$ and defined as ${\rm \textbf{Tr}}(\M \B{N}^\top)$. Given a function $f: \R \rightarrow \R$ defined on scalars, $f(\Q)$ will denote applying the function entry-wise to each entry of the matrix $\Q$.

\section{Proofs for Section \ref{DropBlock}}
{\it {\bf Lemma \ref{lem:DropBlockDet}}{\rm (Restated)} The stochastic DropBlock objective \eqref{eq:DropBlockNN} is equivalent to a regularized deterministic objective:
\begin{align}
&\E_{\w} \left \| \Y - \frac{1}{\theta} \U (\diag{\w} \otimes \I_{\frac{d}{r}}) \V^\top \X \right\|_\F^2 = \| \Y - \U  \V^\top \X \|_\F^2 + \Omega_{\rm DropBlock}(\U, \X^\top \V) \nonumber
\end{align}
where $\Omega_{\rm DropBlock}$ is given by
\begin{equation}
    \Omega_{\rm DropBlock}(\U, \V) = \frac{1 - \theta}{\theta} \sum_{i = 1}^k \| \U_i \V^\top_i \X \|_\F^2 \nonumber
\end{equation}
with $\U_i \in \R^{a\times r}$ and $\V_i \in \R^{b \times r}$ denoting the $i^\text{th}$ blocks of $r$ consecutive columns in $\U$ and $\V$ respectively and $k = \frac{d}{r}$ denoting the number of blocks.
}
\proof This result is an instance of the more general result in Theorem \ref{th:gendet}, and hence we will just specify the Covariance matrix $\C$, mean vector $\bmu$, and mapping $g_\Gamma(\X)$ to be used in order to apply Theorem \ref{th:gendet}.

Recall from the main text that the entries of $\w$ are sampled i.i.d. from ${\rm Ber}(\theta)$. Now, the mean $\bmu$ is given by $\E_\w [ \diag{\w} \otimes \I_{\frac{d}{r}} ] = \theta \I_r \otimes \I_{\frac{d}{r}} = \theta \I$. To compute the covariance matrix, observe that the random variables $z_i, z_j$ are uncorrelated when $i, j$ lie in different blocks. Hence, for such pairs, $c_{i, j} = 0$. Now, when $i, j$ are within the same block ($i$ might equal $j$), either both are dropped with probability $1 - \theta$, or none of them is. Hence, ${\rm Cov}(z_i, z_j) = \E[z_i z_j] - \E[z_i] \E[z_j] = \theta - \theta^2 = \theta (1 - \theta)$. Note that this implies that $\bar \C$ is block diagonal with blocks $\frac{1 - \theta}{\theta} \1_r \1_r^\top$. Finally, we take $g_\Gamma(\X) = \V^\top \X$. Applying Theorem \ref{th:gendet} now, we get the required result. \endproof

{\it {\bf Lemma \ref{lemma:zero}}{\rm (Restated)} Given any matrix $\A$, if the number of columns, $d$, in $(\U,\V)$ is allowed to vary, with $\theta$ held constant, then
\begin{align}
\inf_d \inf_{\substack{\U \in \R^{m\times d},\V \in \R^{n\times d} \\ \A=\U \V^\top \X}} \Omega_{\rm DropBlock}(\U, \X^\top \V) = 0 \nonumber
\end{align}
}
\proof Let $\A = \U \V^\top \X$ be a factorization of $\A$ into $d$ sized factors $\U \in \R^{a \times d}, \V \in \R^{b \times d}$. Construct $\bar \U \in \R^{a \times 2d}, \bar \V \in \R^{b \times 2d}$ by concatenating $\U$ and $\V$ along the $2^{\rm nd}$ axis, as $\bar \U = \frac{1}{\sqrt{2}} \left [\U \ \U \right]$ and  $\bar \V = \frac{1}{\sqrt{2}} \left [\V \ \V \right]$. Observe that $\bar \U \bar{\V}^\top \X = \A$, but 
\begin{align}
\Omega_{\rm DropBlock}(\bar \U, \X^\top \bar \V) &= \sum_{i = 1}^\frac{2d}{r} \|\bar \U_i \bar \V_i ^\top \X \|_\F^2 = 2 \sum_{i = 1}^\frac{d}{r} \left \| \frac{\U_i}{\sqrt{2}} \frac{\V_i^\top \X }{\sqrt{2}} \right \|_\F^2 \nonumber \\ 
&= \frac{1}{2} \Omega_{\rm DropBlock}(\U, \X^\top \V) \nonumber
\end{align}
Hence, by increasing the size of the factorization, we have been able to reduce the regularizer value by half, while maintaining the value of the product. This process can be continued indefinitely to make the value of $\Omega_{\rm DropBlock}$ go arbitrarily close to zero. \endproof

{\it {\bf Theorem \ref{th:eq}}{\rm (Restated)} With $f$ defined as \eqref{fuv}:
\[f(\U, \V, d) = \| \Y - \U  \V^\top \X \|_\F^2  +  \frac{d}{r} \Omega_{\rm DropBlock}(\U, \X^\top \V) \nonumber \]
If $(\U^*, \V^*, d^*)$ is a minimizer of $f$, then $(\U^*, \V^*)$ is balanced. 
}
\begin{proof}
Fix a factorization $(\U, \V, d)$ and suppose $(\U, \V)$ is not balanced. We will then construct a new factorization $(\widehat{\U}, \widehat{\V}, \hat{d} + d)$ such that $\widehat{\U} \widehat{\V}^\top \X = \U  \V^\top \X$, while
  $(d + \hat{d}) \Omega_{\rm DropBlock}(\widehat \U, \X^\top \widehat \V) < d \Omega_{\rm DropBlock}(\U, \X^\top \V)$, showing that $(\U, \V, d)$ is not a global minimizer.

First, define a vector $\balpha \in \R^d$ containing the scale of each factor, $\alpha_i = \|\U_i \V_i^\top \X\|_\F$. Secondly, let $(\bar \U_i, \bar \V_i)$ denote the $i$th normalised factor, $\bar \U_i = (\frac{1}{\alpha_i})^{\frac{1}{2}} \U_i$, $\bar \V_i = (\frac{1}{\alpha_i})^{\frac{1}{2}} \V_i$.

Note that since $(\U, \V)$ is not balanced, we know that $d \Omega_{\rm DropBlock}(\U, \X^\top \V) = d \|\balpha\|_2^2 > \|\balpha\|_1^2$. The last inequality holds because for any $C > 0$, $\frac{C}{d} \1$ is the unique minimum $\ell_2$-norm element over the scaled standard simplex $C \Delta^d$, with $\|\frac{C}{d} \1\|_2^2 = \frac{C^2}{d}$, and we know $\balpha \in \|\balpha\|_1 \Delta^d$.

Now, for a fixed $\hat{d} > 0$, we construct $(\widehat{\U}, \widehat \V)$ of size at most $\hat{d} + d$ by replicating each $(\bar \U_i, \bar \V_i)$ in proportion to $\alpha_i$. Specifically, let $r_i = \lfloor \frac{\alpha_i}{\|\balpha\|_1} \hat{d} \rfloor$ and $\gamma_i = \frac{\alpha_i}{\|\balpha\|_1} \hat{d} - r_i < 1$. Then for each $i = 1, \dotsc, d$, $\widehat{\U}$ contains $r_i$ copies of $(\frac{\|\balpha\|_1}{\hat d})^{\frac{1}{2}} \bar \U_i$ followed by one ``remainder'' factor $(\frac{\gamma_i \|\balpha\|_1}{\hat d})^{\frac{1}{2}} \bar \U_i$. Similarly for $\widehat{\V}$. More explicitly,
\begin{align}
  \widehat{\U} &= 
  \begin{bmatrix}
    \Bigl(\frac{\|\balpha\|_1}{\hat d}\Bigr)^{\frac{1}{2}} \Bigl( [\1_{r_1}^\top \: (\gamma_1)^{\frac{1}{2}}] \otimes \bar \U_1 \Bigr) 
    & \cdots & 
    \Bigl(\frac{\|\balpha\|_1}{\hat d}\Bigr)^{\frac{1}{2}} \Bigl( [\1_{r_d}^\top \: (\gamma_d)^{\frac{1}{2}}] \otimes \bar \U_d \Bigr) 
  \end{bmatrix} \\
  \widehat{\V} &= 
  \begin{bmatrix}
    \Bigl(\frac{\|\balpha\|_1}{\hat d}\Bigr)^{\frac{1}{2}} \Bigl( [\1_{r_1}^\top \: (\gamma_1)^{\frac{1}{2}}] \otimes \bar \V_1 \Bigr) 
    & \cdots & 
    \Bigl(\frac{\|\balpha\|_1}{\hat d}\Bigr)^{\frac{1}{2}} \Bigl( [\1_{r_d}^\top \: (\gamma_d)^{\frac{1}{2}}] \otimes \bar \V_d \Bigr) 
  \end{bmatrix}.
\end{align}

Then observe that $(\widehat{\U}, \widehat{\V})$ are of size $\sum_i r_i + d \leq \hat{d} + d$. And we have by construction
\begin{align}
 \widehat \U \widehat \V^\top \X
  &= \sum_{i=1}^d \frac{\|\balpha \|_1 (r_i + \gamma_i)}{\hat{d}}
  (\bar \U_i \bar \V_i^\top \X)
  = \sum_{i=1}^d \alpha_i (\bar \U_i \bar \V_i^\top \X)
  = \U \V^\top \X \\
  (\hat{d} + d)  \Omega_{\rm DropBlock}(\widehat \U, \X^\top \widehat \V) &= (\hat{d} + d)
  {\Bigl( \frac{\|\balpha\|_1}{\hat{d}} \Bigr)}^2 \sum_{i=1}^d r_i
   + (\hat{d} + d) \sum_{i=1}^d
  {\Bigl(\frac{\gamma_i \|\balpha\|_1}{\hat{d}} \Bigr)}^2 <
  {\Bigl( \frac{\hat{d} + d}{\hat{d}} \Bigr)}^2 \|\balpha\|_1^2.
\end{align}
  Taking $\hat{d}$ sufficiently large so that $\|\balpha\|_1^2 {\Bigl( \frac{\hat{d} +
  d}{\hat{d}} \Bigr)}^2 \leq d \|\balpha\|_2^2 = d \Omega_{\rm DropBlock}(\U, \X^\top \V)$, we see that 
  \[(\hat d + d) \Omega_{\rm DropBlock}(\widehat \U, \X^\top \widehat \V) < d \Omega_{\rm DropBlock}(\U, \X^\top \V)\] and this completes the proof.
\end{proof}

{\it {\bf Theorem \ref{th:tight}}{\rm (Restated)} If $(\U^*, \V^*, d^*)$  is a global minimizer of the factorized problem $f$, then $\A^* = \U^{*} \V^{*\top} \X$ is a global minimizer of the following function $\bar{\bar{ F}}$. 
\begin{equation}
  \bar{\bar{F}}(\A) = \|\Y - \A\|_F^2 + \Theta(\A) \quad \text{where} \quad
  \Theta(\A) \triangleq \inf_{\substack{d, \U \in \R^{m \times d}, \V \in \R^{n
  \times d} \\ \A = \U \V^\top \X}} \Bigl( \sum_{i=1}^k \|\U_i \V_i^\top \X\|_F
  \Bigr)^2
\end{equation}
Futhermore, $\bar{\bar{F}}$ is a convex lower bound to $f$, and is tight at the global minimizer, i.e. we have $f(\U^*, \V^*,d^*) = \bar{\bar{ F}}(\A^*)$. Moreover, the same statements also hold for the lower convex envelope $\F$.
}
\proof Recall that $\bar F$ is defined as:
\begin{equation}
\bar F(\A) = \|\Y-\A \|_F^2 + \frac{1 - \bar \theta}{\bar \theta} \inf_d \inf_{\substack{\U \in \R^{a\times d},\V \in \R^{b\times d} \\ \A=\U \V^\top \X}} \frac{d}{r} \sum_{i = 1}^k \|\U_i \V^\top_i \X \|^2_\F
\end{equation}
Defining the vector $\balpha$ as $\alpha_i = \|\U_i \V^\top_i \X \|_\F$, we see the following (recall that $k = \frac{d}{r}$ and $r$ is constant):
\begin{align}
\|\balpha\|_1^2 &\leq k \|\balpha\|_2^2 \\
\implies \inf_d \inf_{\substack{\U \in \R^{a\times d},\V \in \R^{b\times d} \\ \A=\U \V^\top \X}} \Bigl( \sum_{i=1}^d \|\U_i \V_i^\top \X\|_F
  \Bigr)^2 &\leq  \inf_d \inf_{\substack{\U \in \R^{a\times d},\V \in \R^{b\times d} \\ \A=\U \V^\top \X}} k \sum_{i = 1}^k \|\U_i \V^\top_i \X \|^2_\F \label{thetalambda} \\
\implies \bar{\bar{F}}(\A) &\leq \bar F(\A)
\end{align}

Hence, $\bar{\bar{F}}$ is a lower bound for $\bar F$. Further, we can show that $\Omega(\A)$ is convex. Firstly, we define the function $\theta(\A)$ as 
\[\omega(\A) = \inf_d \inf_{\substack{\U \in \R^{a\times d},\V \in \R^{b\times d} \\ \A=\U \V^\top \X}}  \sum_{i=1}^k \|\U_i \V_i^\top \X\|_F \]
We observe that $\omega$ is a gauge function (as studied in \cite{Haeffele:PAMI2019}), and hence convex and non-negative. Further, we observe that $(\cdot)^2$ is increasing, and convex on $[0, \infty]$. This leads to $(\theta(\alpha \A_1 + (1 - \alpha) \A_2))^2 \leq (\alpha \theta(\A_1) + (1 - \alpha) \theta(\A_2))^2 \leq \alpha (\theta(\A_1))^2 + (1 - \alpha) (\theta(\A_2))^2$. Since $\Theta(\A) = (\theta(\A))^2$, this shows that $\Theta$ is convex. Hence, $\bar{\bar{F}}$ is a convex lower bound for $\bar F$. 

Now, fix $(\U, \V, d)$ and let $\A = \U \V^\top \X$. Suppose $\A$ does not minimize the lower bound $\bar{\bar{F}}$. Then, there exists $\A'$ and $\epsilon > 0$ such that $\bar{\bar{ F}}(\A') \leq \bar {\bar{F}}(\A) - \epsilon$. Now, choose $\U', \V', d'$ such that $\A' = \U' \V'^\top \X$ and $(\sum_i \|\U_i' \V_i'^\top \X\|_F )^2 \leq \Theta(\A') + \epsilon/3$. Apply Theorem \ref{th:eq} to approximately balance and obtain $(\widehat \U, \widehat \V, \hat d + d')$ such that $(\hat d + d') \Omega_{\rm DropBlock}(\widehat \U, \X^\top \widehat \V) < (\sum_i \|\U_i' \V_i'^\top \X\|_F )^2 + \epsilon/3$. It follows that:
\[ f(\widehat \U, \widehat \V, \hat d + d') < \bar{\bar{ F}}(\A') + 2\epsilon/3 \leq \bar{\bar{ F}}(\A) - \epsilon/3 \leq f(\U, \V, d) - \epsilon / 3, \]
showing that $(\U, \V, d)$ is not a global minimizer of $f$. 

So, we have shown that for any global minimizer $(\U^*, \V^*, d^*)$ of the factorized objective $f$, $\A^* = \U^* \V^{*\top} \X$ must also be a global minimizer of $\bar{\bar{F}}$. Since by Theorem \ref{th:eq}, $(\U^*, \V^*, d^*)$ must be balanced, we know the objectives must also be equal, $\bar{\bar{F}}(\A^*) = f(\U^*, \V^*, d^*)$.

We have seen that $\bar{\bar{F}}$ is a convex lower bound for $\bar F$. However, we know that $\F$ is the lower convex envelope of $\bar{F}$, that has the highest value at each point $\A$ among all convex lower bounds to $\bar F$. This means that $\bar{\bar{F}}(\A) \leq \F(\A)$ for all $\A$. Further, $\bar{F}$ lower bounds $f$, and hence we must have $\bar{\bar{F}}(\A) \leq F(\A) \leq \bar{F}(\A) \leq f(\U, \V, d)$ for all $\A$ and $(\U, \V, d)$ such that $\A = \U \V^\top \X$. Due to this chain of inequalities, and the fact that we have shown $\bar{\bar{F}}(\A^*) = f(\U^*, \V^*, d^*)$, it implies that all of the functions are equal at $\A^*$. Hence, what we have shown for $\bar{\bar{F}}$ must also hold for $\F$.
\endproof

{\it {\bf Corollary \ref{cor:lambdaconvex}}{\rm (Restated)} $\Lambda(\A)$ is convex and equal to its lower convex envelope $\Lambda^{**}(\A)$. 
\proof Recall the definition of $\Lambda(\A)$, and that $k = \frac{d}{r}$ where $r$ is constant: 
\begin{equation}  \Lambda(\A) = \inf_d \inf_{\substack{\U \in \R^{a\times d},\V \in \R^{b\times d} \\ \A=\U \V^\top \X}} k \sum_{i = 1}^k \|\U_i \V^\top_i \X \|^2_\F \label{lambdainf} \end{equation}
As in the proof of Theorem \eqref{th:tight}, define $\Theta(\A)$ as:
\begin{equation} \Theta(\A) = \inf_d \inf_{\substack{\U \in \R^{a\times d},\V \in \R^{b\times d} \\ \A=\U \V^\top \X}}  \Big( \sum_{i = 1}^k \|\U_i \V^\top_i \X \|_\F \Big)^2 \label{thetainf} \end{equation}
Again, from Eq. \eqref{thetalambda}, observe that $\Theta(\A) \leq \Lambda(\A)$ for all $\A$. Assume that there exists $\A$ such that $\Theta(\A) + \epsilon \leq \Lambda(\A)$ for some $\epsilon > 0$. Let $(\U', \V', d')$ be an $(\epsilon/3)$ optimal point for $\Theta(\A)$ such that $\Big( \sum_{i = 1}^{k'} \|\U'_i {\V'}^\top_i \X \|_\F \Big)^2 \leq \Theta(\A) + (\epsilon/3)$, and $\U' {\V'}^\top \X = \A$. Apply Theorem \ref{th:eq} to obtain $(\widehat \U, \widehat \V, \widehat d)$ such that the following holds ($\widehat k = \frac{\widehat d}{r}$ and $k' = \frac{d'}{r}$):
\begin{align} 
\widehat k \sum_{i = 1}^{\widehat k} \|\widehat \U_i {\widehat{\V}}^\top_i \X \|^2_\F &\leq 
\Big( \sum_{i = 1}^{k'} \|\U'_i {\V'}^\top_i \X \|_\F \Big)^2 + \frac{\epsilon}{3} \\
\implies \widehat k \sum_{i = 1}^{\widehat k} \|\widehat \U_i {\widehat{\V}}^\top_i \X \|^2_\F &\leq 
\Theta(\A) + \frac{2\epsilon}{3} \label{balancingresult}
\end{align}
But since $\widehat \U {\widehat \V}^\top \X = \A$ (the balancing procedure does not affect the product), the LHS of \eqref{balancingresult} is considered in the infimum of $\Lambda(\A)$ \eqref{lambdainf}. Hence, $\Lambda(\A) \leq \Theta(\A) + 2\epsilon/3 < \Theta(\A) + \epsilon \leq \Lambda(\A)$, which is a contradiction. Hence, we have $\Theta(\A) = \Lambda(\A)$. Further, since $\Theta$ is convex (as shown in the proof of Theorem \ref{th:tight}), we have $\Theta(\A) \leq \Lambda^{**}(\A) \leq \Lambda(\A)$. But since $\Theta(\A) = \Lambda(\A)$, we have shown that $\Lambda^{**}(\A) = \Lambda(\A)$.

{\it {\bf Theorem \ref{th:convexlb}}{\rm (Restated)} The lower convex envelope of the DropBlock regularizer $\Lambda(\A)$ in \eqref{dropblockreg}, is given by
\begin{equation}
\Lambda^{**}(\A) = \frac{1 - \bar \theta}{\bar \theta} \left( \sum_{i = 1}^{\rho^* - 1} a_i^2 + \frac{(\sum_{i = \rho^*}^d a_i)^2}{r - \rho^* + 1} \right), \nonumber
\end{equation}
where $\rho^*$ is the integer in $\{1, 2, \ldots, r\}$ that maximizes \eqref{convexlb}, and $a_1 \geq a_2 \ldots \geq a_d$ are the singular values of $\A$.
}
\proof We begin by noting that if the lower convex envelope of a function $f(x)$ is given by $g(x)$, then the lower convex envelope of $\alpha f(x)$ is given by $\alpha g(x)$ for a constant $\alpha \geq 0$. We will use this fact to simplify the presentation of the following proof, by finding the lower convex envelope of $\bar \Lambda(\A)$ such that $\Lambda(\A) = \frac{1 - \bar \theta}{\bar{\theta}} \bar \Lambda(\A)$, since $\bar \theta$ is a constant. Hence, $\bar \Lambda(\A)$ is defined as,
\begin{equation}
    \bar \Lambda(\B{A}) = \inf_{\substack{\U \in \R^{m\times d},\V \in \R^{n\times d},d \\ \B{A}=\U \V^\top \X}} \frac{d}{r} \sum_{i = 1}^k \|\B{U}_i \B{V}^\top_i \X\|^2_F
\end{equation}

Computing the Fenchel-Conjugate of $\bar \Lambda(\A)$, 
\begin{align}
    \bar \Lambda^*(\Q) 
    &= \sup_\A \left( \langle \Q, \A \rangle - \inf_{\substack{\U \in \R^{m\times d},\V \in \R^{n\times d},d \\ \B{A}=\U \V^\top \X}} \frac{d}{r} \sum_{i = 1}^k \|\B{U}_i \B{V}^\top_i \X\|^2_F \right) \nonumber \\
    &= \sup_{\U, \V, d} \left( \langle \Q, \U \V^\top \X \rangle - \frac{d}{r} \sum_{i = 1}^k \|\B{U}_i \B{V}^\top_i \X\|^2_F \right) \nonumber \\
    &= \sup_{\U, \V, d} \left( \langle \Q, \sum_{i = 1}^k \B{U}_i \B{V}^\top_i \X \rangle - \frac{d}{r} \sum_{i = 1}^k \|\B{U}_i \B{V}^\top_i \X \|^2_F \right) \nonumber \\
    &= \sum_{i = 1}^k \sup_{\U_i, \V_i, d} \left( \langle \Q, \B{U}_i \B{V}^\top_i \X \rangle - \frac{d}{r} \|\B{U}_i \B{V}^\top_i \X \|^2_F \right) \label{supSeparable}
\end{align}
In the above, (\ref{supSeparable}) follows because we can separate the maximization over the pairs $(\U_i, \V_i)$. The value of all these maximization problems are the same. Let $\U_i \V_i^\top \X = \M \Sigma \B{N}^\top$ be a singular value decomposition of the product $\U_i \V_i^\top \X$, with $\Sigma = \diag{\sigma}$. Since $\U_i, \V_i$ each have $r$ columns, we know that the number of non-zero entries in $\sigma$ is atmost $r$, i.e. $\|\sigma\|_0 \leq r$, hence the problem (\ref{supSeparable}) now reduces to:
\begin{align}
   \bar \Lambda^*(\Q)
   &= k \cdot \sup_{\|\sigma\|_0 \leq r, d, \B{M}, \B{N}} \left( \langle \Q, \B{M} \diag{\sigma} \B{N}^\top \rangle - \frac{d}{r} \|\B{M} \diag{\sigma} \B{N}^\top\|^2_F \right) \nonumber \\
   &= k \cdot \sup_{\|\sigma\|_0 \leq r, d, \B{M}, \B{N}} \left( \langle \diag{\sigma_Q}, \diag{\sigma} \rangle - \frac{d}{r} \|\sigma\|_2^2 \right) \label{vonneumann} \\
   &= k \cdot \sup_{\|\sigma\|_0 \leq r, d} \sigma_Q^\top \sigma - \frac{d}{r} \|\sigma\|^2_2 \nonumber \\
   &= \frac{1}{4} \sum_{i = 1}^{r} \sigma_i^2(\Q) \label{dkr}
\end{align}
In the above, (\ref{vonneumann}) follows from the Von-Neumann trace inequality, when $\Q = \B{M_Q} \diag{\sigma_Q} \B{N_Q}^\top$, and $\sigma$ has the same ordering as $\sigma_Q$ and $\B{M}^\top \B{M_Q} = \B{N_Q}^\top \B{N} = {\rm I_d} $. To obtain \eqref{dkr}, we use $d = k \cdot r$. Note that there is an implicit sufficient rank assumption for obtaining \eqref{dkr}, i.e. there should be a choice of $\U_i \V_i$ such that the top $r$ singular values of $\U_i \V_i^\top \X$ are the same as $\sigma_Q$. We compute the double dual now. 

The basic geometric idea behind the computation of the double dual is to understand the shape of the constraint set, and the level sets of the objective function \eqref{dkr}. The constraint set $x_1 \geq x_2 \geq \ldots x_d$ is a cone in $d$ dimensions.
\begin{align}
   \bar \Lambda^{**}(\A) &= \sup_{\X} \left( \langle \A, \X \rangle - \frac{1}{4} \sum_{i = 1}^{r} \sigma_i^2(\X) \right) \label{neumannagain} \\
   &= \sup_x \left( \sum_{i = 1}^{d} a_i x_i - \frac{1}{4} \sum_{i = 1}^{r} x_i^2 \right) \text{ s.t. } x_1 \geq x_2 \geq \ldots \geq x_d \geq 0 \label{notationhere}
\end{align}
To simplify notation, in (\ref{notationhere}), we denote the singular values of $\A$ by $a_i$ and the singular values of $\X$ by $x_i$, with $a_1 \geq a_2 \geq \ldots \geq a_d$. Note that we have used the Von-Neumann inequality to obtain \eqref{notationhere} from \eqref{neumannagain}, and hence, the SVD of $\X$ is given by $\U_\A \diag{\x_{\rm sol}} \V_\A^\top$ where $\A = \U_\A \diag{\a} \V_\A^\top$ is the SVD of $\A$.

To study (\ref{notationhere}), we first note that the variables $x_{r + 1}, \ldots, x_d$ are bounded only by the order constraints and non-negativity constraints on $x_i$. Since all the $a_i \geq 0$, the objective \eqref{notationhere} is linear and monotonously increasing in any of the variables $x_{r + 1}, \ldots, x_d$. Hence, in any optimal solution to (\ref{notationhere}), each of $x_{r:d}$ takes the maximum value possible within constraints, and we have $x_{r} = x_{r + 1} = \ldots = x_d$ at optimality. We have the following simplified problem now:
\begin{align}
    &\sup_x \left( \sum_{i = 1}^{r - 1} a_i x_i + x_r \sum_{i = r}^{d} a_i - \frac{1}{4} \sum_{i = 1}^{r} x_i^2 \right) \label{constraintdrop} \\
    &\quad \text{ s.t. } x_1 \geq x_2 \geq \ldots \geq x_r \geq 0
\end{align}
Notice that the solution to the optimization problem \eqref{constraintdrop} does not change if we drop the constraints among the first $r - 1$ variables. Hence, \eqref{constraintdrop} is same as:
\begin{align}
    &\sup_x \left( \sum_{i = 1}^{r - 1} a_i x_i + x_r \sum_{i = r}^{d} a_i - \frac{1}{4} \sum_{i = 1}^{r} x_i^2 \right) \nonumber \\
    &\quad \text{ s.t. } \min (x_1, x_2, \ldots, x_r) = x_r \geq 0
\end{align}
Now, assume that the minimum among $x_i$ is $x_\rho$. We consider two cases, $\rho = 1$, and $\rho > 1$. For the case when $\rho = 1$, or when all of the $x_i$ are equal, we see that the optimization problem is:
\begin{align}
    \sup_{x_1} \left( x_1 \sum_{i = 1}^{d} a_i - \frac{r}{4} x_1^2 \right) \text{ s.t. } x_1 \geq 0
\end{align}
This has the solution $x_1 = \frac{2}{r} \sum_{i = 1}^d a_i$. 

Now, consider the case when $\rho > 1$. Recall that $\rho \leq r$. WLOG, let the solution vector $x$ have the form $(x_1, x_2, x_3, \ldots, x_\rho, x_\rho, \ldots, x_\rho)$, with $x_{\rho - 1} > x_\rho$. 

\textbf{Claim 1:} We claim that $x_i = 2 a_i$ for $i = 1, 2, \ldots, \rho - 1$. Since there is no upper constraint on $x_1$, and $x_1$ contributes the factor $a_1 x_1 - \frac{1}{4} x_1^2$, we see that $x_1 = 2 a_1$ at optimality. Now, assume inductively that the claim is true for $x_1, \ldots, x_i$. Consider $x_{i + 1}$. Since $x_i \geq x_{i + 1}$ and $x_{i} = 2 a_i$, we know that the constraint on $x_{i + 1}$ is $0 \leq x_{i + 1} \leq 2 a_i$. 

Observe that $x_{i + 1}$ contributes the factor $a_{i + 1} x_{i + 1} - \frac{1}{4} x_{i + 1}^2$ to the sum, which has the maximum at $x_{i + 1} = 2 a_{i + 1}$. As $0 \leq 2 a_{i + 1} \leq 2 a_i$, this lies in the permissible range for $x_{i + 1}$, and we see that $x_{i + 1} = 2 a_{i + 1}$. This proves the claim via induction. 

\textbf{Claim 2:} We claim that $x_\rho = 2 \frac{\sum_{i = \rho}^d a_i}{r - \rho + 1}$. Observe that $x_{\rho}$ contributes the factor $x_\rho \sum_{i = \rho}^d a_i - \frac{1}{4} (r - \rho + 1) x_{\rho}^2$ to the sum, which has the maximum at $\alpha = \frac{2}{r - \rho + 1} \sum_{i = \rho}^d a_i$. Along with the permissible range $0 \leq x_\rho \leq 2 a_{\rho - 1}$, we see that if $\alpha > 2 a_{\rho - 1}$, then the maximum occurs at $x_\rho = 2 a_{\rho - 1} = x_{\rho - 1}$, which violates the assumption $x_{\rho - 1} > x_\rho$. Hence, $\alpha \leq 2 a_{\rho - 1}$, and $x_\rho = \alpha$. 

To summarise, given $\rho$, the solution x has the following closed form:
\begin{align}
&\left( 2 a_1, 2 a_2, \ldots, 2 a_{\rho - 1}, \frac{2}{r - \rho + 1} \sum_{i = \rho}^d a_i, \ldots, \frac{2}{r - \rho + 1} \sum_{i = \rho}^d a_i \right) \label{soln}
\end{align}

The objective value at the solution (\ref{soln}) is:
\begin{align}
    &\sum_{i = 1}^{r - 1} a_i x_i + x_r \sum_{i = r}^{d} a_i - \frac{1}{4} \sum_{i = 1}^{r} x_i^2 \nonumber \\
    &= \sum_{i = 1}^{\rho - 1} (a_i x_i - \frac{1}{4} x_i^2) + x_\rho (\sum_{i=\rho}^d a_i) - \frac{r - \rho + 1}{4} a_\rho^2 \nonumber \\
    &= \sum_{i = 1}^{\rho - 1} a_i^2 + \frac{2 \sum_{i = \rho}^d a_i}{ r - \rho + 1} (\sum_{i=\rho}^d a_i) - (r - \rho + 1) (\frac{\sum_{i = \rho}^d a_i}{ r - \rho + 1})^2 \nonumber \\
    &= \sum_{i = 1}^{\rho - 1} a_i^2 + \frac{(\sum_{i = \rho}^d a_i)^2}{ r - \rho + 1} \label{finalsol}
\end{align} 
To find the value of $\rho$, it suffices to evaluate $\rho = \{1, 2, \ldots, d\}$ to get the value $\rho^*$ that maximizes \eqref{finalsol}. Finally, bringing the multiplier back, we have the following expression for $\Lambda^{**}(\A)$:
\begin{align}
\Lambda^{**}(\A) &= \frac{1 - \bar \theta}{\bar \theta} \bar \Lambda^{**}(\A) \nonumber \\ 
&= \frac{1 - \bar \theta}{\bar \theta} \left( \sum_{i = 1}^{\rho^* - 1} a_i^2 + \frac{(\sum_{i = \rho^*}^d a_i)^2}{r - \rho^* + 1} \right) \nonumber 
\end{align} \endproof

{\it {\bf Theorem \ref{th:globmin}}{\rm (Restated)} The global minimizer of $\F(\A)$ is given by $\A_{\rho, \lambda} = \U_\Y \diag{\a_{\rho, \lambda}} \V_\Y^\top$, where $\Y = \U_\Y \diag{\m} \V_\Y^\top$ is an SVD of $\Y$, and $\a_{\rho, \lambda}$ is given by%
\begin{equation}
\a_{\rho, \lambda}=
\begin{cases}
\left( \frac{m_1}{\beta + 1}, \frac{m_2}{\beta + 1}, \ldots, \frac{m_\lambda}{\beta + 1}, 0, \ldots, 0 \right) 
& \text{if } \lambda \leq \rho - 1\\
\left(
\begin{array}{ll}
\frac{m_1}{\beta + 1}, \frac{m_2}{\beta + 1}, \ldots, \frac{m_{\rho - 1}}{\beta + 1}, \\
m_{\rho} - \frac{\beta}{c} S, m_{\rho + 1} - \frac{\beta}{c} S, \ldots, \\
m_{\lambda} - \frac{\beta}{c} S, 0, \ldots, 0
\end{array}
\right.
& \text{if } \lambda \geq \rho. 
\end{cases} \nonumber
\end{equation}

The constants are $\beta = \frac{1 - \bar \theta}{\bar \theta}$, $S = \sum_{i = \rho}^\lambda m_i$, $c = r + \beta \lambda + (\beta + 1) (1 - \rho)$. $\rho \in \{1, 2, \ldots, r - 1\}$ and $\lambda \in \{1, 2, \ldots, d\}$ are chosen such that they minimize $\F(\A_{\rho, \lambda})$.
}
\proof We now use the solution for $\Lambda^{**}$ computed above to find the global minimizer of the convex lower bound defined in \eqref{lowerbd} as $\F(\A) = \|\M - \A\|_\F^2 + \Lambda^{**}(\A) = \|\M - \A\|_\F^2 + \beta \bar \Lambda^{**}(\A)$, with $\beta = \frac{1 - \bar \theta}{\bar \theta}$. 
\begin{align}
    & \min_\X \|\A - \X\|_F^2 + \beta \bar \Lambda^{**}(\X) \nonumber \\
    &= \min_\X \|\sigma_\A - \sigma_\X\|^2_F + \beta \sum_{i = 1}^{\rho - 1} x_i^2 + \beta \frac{(\sum_{i = \rho}^d x_i)^2}{(r - \rho + 1)} \label{vn} \\
    &= \min_{\B{x}} \sum_{i = 1}^d(a_i - x_i)^2 + \beta \sum_{i = 1}^{\rho - 1} x_i^2 + \beta \frac{(\sum_{i = \rho}^d x_i)^2}{(r - \rho + 1)} \label{reducedopt}
\end{align}

Note that in moving from (\ref{vn}) to (\ref{reducedopt}), we used the Von-Neumann inequality, and took $\X = \U_\A \diag{\x} \V_\A^\top$ where $\A = \U_\A \diag{\a} \V_\A^\top$ is an SVD of $\A$. Assume there are $\lambda$ non-zero elements in the solution of \eqref{reducedopt}. The objective now becomes:
\begin{align}
    \min_{\B{x}} \sum_{i = 1}^\lambda(a_i - x_i)^2 + \sum_{i=\lambda+1}^d a_i^2 + \beta \sum_{i = 1}^{\lambda} x_i^2 \label{reducedopt2}
\end{align}

From the KKT conditions for (\ref{reducedopt2}), we have $x_1, x_2, \ldots, x_\lambda > 0$ and $x_\lambda = x_{\lambda + 1} = \ldots = x_d = 0$. We consider two cases, $\lambda \leq \rho$ and $\lambda > \rho$. Firstly, consider the case when $\lambda \leq \rho - 1$. From the KKT conditions we have the following solution for this case:
\begin{align}
2(x_i - a_i) + 2 \beta x_i = 0 \implies x_{1:\lambda} = \frac{1}{\beta + 1} a_{1:\lambda} \label{sol1}
\end{align}

For $\lambda > \rho$, the objective becomes
\begin{align}
    \min_{\B{x}} \sum_{i = 1}^\lambda(a_i - x_i)^2 + \sum_{i=\lambda+1}^d a_i^2 + \beta \sum_{i = 1}^{\rho - 1} x_i^2 + \beta \frac{(\sum_{i = \rho}^\lambda x_i)^2}{(r - \rho + 1)} \label{reducedopt3}
\end{align}

For the first $\rho - 1$ variables, $x_{1:\rho - 1}$, the KKT conditions give the same solution as in (\ref{sol1}), i.e. $x_{1:\rho - 1} = \frac{1}{\beta + 1} a_{1:\rho - 1}$. For $x_i$ with $\rho \leq i \leq \lambda$, we need to solve a system of linear equations, as follows:
\begin{align} 
&2(x_i - a_i) + 2 \beta \frac{\sum_{i = \rho}^d x_i}{(r - \rho + 1)} = 0 \nonumber \\ 
&\implies \left( I_{\lambda - \rho + 1} + \frac{\beta}{(r - \rho + 1)}\B{1} \B{1}^\top \right) x_{\rho:\lambda} = a_{\rho:\lambda} \nonumber\\
&\implies x_{\rho:\lambda} = \left( \left( I_{\lambda - \rho + 1} - \frac{\beta}{(r - \rho + 1) + \beta(\lambda - \rho + 1)}\B{1} \B{1}^\top \right)a_{\rho:\lambda}\right)_+ \label{woodbury} \\
&\implies x_{\rho:\lambda} = \left(a_{\rho:\lambda} - \frac{\beta \sum_{i = \rho}^\lambda a_i}{r + \beta \lambda - (\beta + 1) \rho + (\beta + 1)} \right)_+ \label{sol2}
\end{align}
For (\ref{woodbury}), we use the Sherman-Morrison Matrix Identity i.e. $(I_d + \alpha \B{1 1}^\top)^{-1} = I_d - \frac{\alpha}{1 + \alpha d} \B{1 1}^\top$. The $+$ subscript on \eqref{sol2} denotes applying the thresholding operation $x_i = {\rm Max}(x_i, 0)$ to all entries of the vector.

Note that we do not know $\lambda$ or $\rho$ at this point. However, from the solution (\ref{sol1}, \ref{sol2}), we know that the optimal pair $(\lambda^*, \rho^*)$ achieves the minimum value of $\F(\X)$, given the solution $\X = \U_\A \diag{\x} \V_\A^\top$. Hence, to compute this optimal pair, we can evaluate all $O(d^2)$ possibilities for $(\lambda, \rho)$ given $\A$. \endproof

{\it {\bf Corollary \ref{cor:allequal}}{\rm (Restated)}
If $\A^*$ is a global minimizer of the lower convex envelope $\F$, and $(\U^*, \V^*, d^*)$ is a global minimizer of the non-convex objective $f$, then we have $\F(\A^*) = \bar F(\A^*) = f(\U^*, \V^*, d^*)$ with $\A^* = \U (\V^*)^\top \X$. 
}
\proof By Theorem \ref{th:eq}, we know that if $(\U^*, \V^*, d^*)$ is a global minimizer of $f$, then with $\A^* = \U (\V^*)^\top \X$, we have $\bar F(\A^*) = f(\U^*, \V^*, d^*)$. Additionally, we know that $\A^*$ is a global minimizer of $\bar F$. Now, by construction, we know that $\F$ is the lower convex envelope of $\bar F$. Also, $\bar F$ is a non-negative function defined over the space of real matrices. Hence, by the properties of the lower convex envelope \cite{Rockafellar:2015}, we know that $F$ and $\bar F$ have the same value at $\A^*$. \endproof

\section{Proofs for Section \ref{generalised}}
{\it {\bf Theorem \ref{th:gendet}} {\rm (Restated)} The Generalized Dropout objective (\ref{dropoutNNgeneral}) is equivalent to a regularized deterministic objective:
\begin{align}
&\E_{\z} \left \| \Y - \U \diag{\boldsymbol{\mu}}^{-1} \diag{\z} g_\Gamma(\X) \right\|_\F^2 = \left \| \Y - \U g_\Gamma(\X) \right\|_\F^2 + \Omega_{\C, \bmu}(\U, g_\Gamma(\X)^\top) \nonumber
\end{align}
where the \emph{generalized Dropout} regularizer, $\Omega_{\C,\bmu} (\U, \V)$ is:
\begin{align}
    \Omega_{\C, \bmu} (\U,\V) &= \sum_{i, j = 1}^d c_{i, j} \frac{(\u_i^\top \u_j) (\v_i^\top \v_j)}{\mu_i \mu_j} = \langle \bar \C, \U^\top \U \odot \V^\top \V \rangle ,\nonumber
\end{align}
}
\proof We generalize the Dropout analysis to get a deterministic equivalent for the objective when $\z$ is drawn from an arbitrary distribution $\mathcal{S}$.
\begin{align}
    &\E_{\z} \left \| \Y - \U \diag{\boldsymbol{\mu}}^{-1} \diag{\z} g_\theta(\X) \right\|_\F^2 \nonumber \\
    &\quad = \left \| \Y - \U \diag{\boldsymbol{\mu}}^{-1} \diag{\z} \V^\top \right\|_\F^2 \label{changevar} \\
    &\quad = \left \| \E_{\z} \left [ \Y - \U \diag{\boldsymbol{\mu}}^{-1} \diag{\z}\V^\top \right ] \right\|_\F^2 + \1^\top \Var \left [ \Y - \U \diag{\boldsymbol{\mu}}^{-1} \diag{\z}\V^\top \right ] \1 \label{expectation2} \\
    &\quad = \left \| \Y - \U \diag{\boldsymbol{\mu}}^{-1} \E_{\z} \left [\diag{\z}\right ]\V^\top \right\|_\F^2 + \1^\top \Var \left [\U \diag{\boldsymbol{\mu}}^{-1} \diag{\z}\V^\top \right ] \1 \nonumber \\
    &\quad = \left \| \Y - \U \diag{\boldsymbol{\mu}}^{-1} \diag{\bmu} \V^\top \right\|_\F^2 + \1^\top \Var \left [\sum_{i = 1}^d \frac{z_i}{\mu_i} \u_i \v_i^\top \right ] \1 \nonumber \\
    &\quad = \left \| \Y - \U \V^\top \right\|_\F^2 + \sum_{i, j = 1}^d \1^\top {\rm \bf Covar} \left ( \frac{z_i}{\mu_i} \u_i \v_i^\top, \frac{z_j}{\mu_j} \u_j \v_j^\top \right ) \1 \label{varexpand} \\
    &\quad = \left \| \Y - \U \V^\top \right\|_\F^2 + \sum_{i, j = 1}^d \frac{\langle \u_i \v_i^\top, \u_j \v_j^\top \rangle}{\mu_i \mu_j} {\rm \bf Covar} (z_i z_j) \nonumber \\
    &\quad = \left \| \Y - \U \V^\top \right\|_\F^2 + \sum_{i, j = 1}^d c_{i, j} \frac{\langle \u_i \v_i^\top, \u_j \v_j^\top \rangle}{\mu_i \mu_j} \nonumber
\end{align}
%
%
%
%
%
%
%
%
In \eqref{changevar}, we have substituted $g_\theta(\X) = \V^\top$ for ease of presentation. Then, in (\ref{expectation2}), we have used the identity $\E[a^2] = \E[a]^2 + \Var[a]$ applied to each element of the matrix $\A$, which gives that $\E[ \|\A\|_\F^2 ] = \|\E[\A] \|_\F^2 + \1^\top \Var[\A] \1$. We have used the matrix equivalent of the identity $\Var(a_1 + a_2 + \ldots + a_d) = \sum_{i, j} {\rm \bf Covar} (a_i, a_j)$ in (\ref{varexpand}). \endproof

{\it {\bf Corollary \ref{cor:dropoutextend}}{\rm (Restated)}
For regular Dropout applied to objective \eqref{dropoutNNgeneral} the following equivalence holds:
\begin{align} \label{dropoutDNNreg}
&\E_{\z} \left \| \Y - \U \diag{\boldsymbol{\mu}}^{-1} \diag{\z} g_\Gamma(\X) \right\|_\F^2 = \left \| \Y - \U g_\Gamma(\X) \right\|_\F^2 + \sum_{i=1}^d \|\u_i\|_2^2 \|g_\Gamma^i(\X)\|_2^2
\end{align}
where $g_\Gamma^i(\X) \in \R^N$ denotes the output of the $i^\text{th}$ neuron of $g_\Gamma$ (i.e., the $i^\text{th}$ row of $g_\Gamma(\X)$). 
}
\proof For regular Dropout, we have $\z \overset{\text{i.i.d}}{\sim}$ Bernoulli$(\theta)$. Hence, the covariance matrix is given by $c_{i, i} = \theta (1 - \theta)$ and $c_{i, j} = 0$ when $i \neq j$. Further, $\mu_i = \theta$ for all $i$. Using Theorem \ref{th:gendet} with this choice of $\bmu, \C$ we get the result.

{\it {\bf Proposition \ref{thm:nucnorm_limit}}{\rm (Restated)}
If the network architecture, $g_\Gamma$, has sufficient capacity to span $\R^{d \times N}$ (i.e., given any matrix $\Q \in \R^{d \times N}, \ \exists \bar \Gamma$ such that $g_{\bar \Gamma}(\X) = \Q$) and $d \geq \min\{a,N\}$, then the global optimum of \eqref{dropoutNNgeneral} with $\z \overset{\text{i.i.d}}{\sim}$ Bernoulli$(\theta)$ is given by:
\begin{align}
&\min_{\U,\Gamma} \E_{\z} \left \| \Y - \U \diag{\boldsymbol{\mu}}^{-1} \diag{\z} g_\Gamma(\X) \right\|_\F^2  = \min_{\A} \|\Y - \A\|_F^2 + \tfrac{1-\theta}{\theta} \|\A\|_*^2
\end{align}
where $\|\A\|_*$ denotes the nuclear norm of $\A$.
}
\proof From Corollary \ref{cor:dropoutextend}, we have for Dropout, 
\[ \E_{\z} \left \| \Y - \U \diag{\boldsymbol{\mu}}^{-1} \diag{\z} g_\Gamma(\X) \right\|_\F^2 = \left \| \Y - \U g_\Gamma(\X) \right\|_\F^2 + \sum_{i=1}^d \|\u_i\|_2^2 \|g_\Gamma^i(\X)\|_2^2 \]
Since the network is sufficiently overparameterized such that for any $\V \in \R^{N \times d}$ there exists $\bar \Gamma$ such that $g_{\bar \Gamma}(\X) = \V^\top$, we can replace $g_{\bar \Gamma}(\X)$ by $\V$ and optimize over $\V$. Now, we can use arguments similar to \cite{Mianjy:ICML18} to obtain the result:
\begin{align}
\min_{\U, \V}  \|\Y - \U \V^\top\|_\F^2 + \sum_{i = 1}^{d} \|\u_i\|^2 \|\v_i\|^2 = \min_\A \|\Y - \A\|_\F^2 + \|\A\|_*^2
\end{align} \endproof

\section{Proofs for Section \ref{dropconnect}}
{\it {\bf Theorem \ref{th:dropconnectdnn}} {\rm (Restated)}
For Dropconnect applied to the second-last layer weights $\V$ of a deep network parameterized as $\U \V^\top g_\Gamma(\X)$, the following equivalence holds:
\begin{align*}
&\E_{\Z} \left \| \Y - \frac{1}{\theta} \U(\Z \odot \V)^\top g_\Gamma(\X) \right\|_\F^2 = \left \| \Y - \U \V^\top g_\Gamma(\X) \right\|_\F^2 + \frac{1 - \theta}{\theta}\sum_{i = 1}^{d} \|\u_i\|_2^2 \|g_\Gamma(\X)^\top \v_i\|_2^2
\end{align*}
where $g_\Gamma^i(\X) \in \R^N$ denotes the output of the $i^\text{th}$ neuron of $g_\Gamma$ (i.e., the $i^\text{th}$ row of $g_\Gamma(\X)$). 
}
\proof We show that DropConnect induces a deterministic objective. The proof performs algebraic manipulations to evaluate the expectation of the objective over $\Z$ first and then $\x$. For ease of presentation, we substitute $\M = g_\Gamma(\X)$, and hence evaluate $\mathbb{E}_{\mathbf{Z}} \left \| \Y - \frac{1}{\theta} \U(\mathbf{Z} \odot \V)^\top \M \right\|_\F^2$. 
\begin{align}
    &\mathbb{E}_{\mathbf{Z}} \left \| \Y - \frac{1}{\theta} \U(\mathbf{Z} \odot \V)^\top \M \right\|_\F^2 \label{assumption} \\
    &= \left \| \mathbb{E}_\mathbf{Z} \left[ \Y - \frac{1}{\theta} \U(\mathbf{Z} \odot \V)^\top \M \right] \right\|_\F^2 + \mathbf{1}^\top \Var_\Z{\left[ \Y - \frac{1}{\theta} \U(\mathbf{Z} \odot \V)^\top \M \right] \1} \nonumber \\
    &= \left \| \Y - \frac{1}{\theta} \U\mathbb{E}_\B{Z}(\B{Z} \odot \V)^\top \M \right\|_\F^2 + \frac{1}{\theta^2} \mathbf{1}^\top \Var_\Z{\left[ \U(\mathbf{Z} \odot \V)^\top \M \right] \1} \label{expectation} \\
    &= \left \| \Y - \frac{1}{\theta} \U (\theta \V)^\top \M \right\|_\F^2 + \frac{1}{\theta^2} \mathbf{1}^\top \Var_\Z{\left[ \sum_{k = 1}^{d} \B{u}_k \left(\M^\top (\B{Z} \odot \V) \right)_k \right] \1} \nonumber \\
    &= \left \| \Y - \frac{1}{\theta} \U (\theta \V)^\top \M \right\|_\F^2 + \frac{1}{\theta^2} \mathbf{1}^\top \Var_\Z{\left[ \sum_{k = 1}^{d} \B{u}_k \M^\top (\B{Z} \odot \V)_k \right] \1} \label{zcolIndepPrev2}\\
    &= \left \| \Y -  \U \V^\top \M \right\|_\F^2 + \frac{1}{\theta^2} \sum_{k = 1}^{d} \mathbf{1}^\top \Var_{\z_k}{\left[ \B{u}_k \M^\top (\z_k \odot \B{v}_k) \right] \1}  \label{zcolIndep2} \\
    &= \left \| \Y -  \U \V^\top \M \right\|_\F^2 + \frac{1}{\theta^2} \sum_{k = 1}^{d} (\theta)(1 - \theta) \| \B{u}_k \|_\F^2 \|\M^\top \v_k\|_\F^2 \label{zIndep2} \\
    &= \left \| \Y -  \U \V^\top \M \right\|_\F^2 + \frac{1 - \theta}{\theta} \sum_{k = 1}^{d} ||\B{u}_k||_\F^2 ||\M^\top \B{v}_k||^2_2 \label{xxtcondition}
\end{align}
In (\ref{expectation}), we have used the identity $\E[a^2] = \E[a]^2 + \Var[a]$ applied to each element of the matrix $\A = \Y - \frac{1}{\theta} \U(\mathbf{Z} \odot \V)^\top \M$, which gives that $\E[ \|\A\|_\F^2 ] = \|\E[\A] \|_\F^2 + \1^\top \Var[\A]$. Each of the columns of $\B{Z}$ are independent, which implies each of the summands in (\ref{zcolIndepPrev2}) are independent. Hence, the variance and the summation can be interchanged, and (\ref{zcolIndep2}) follows. 
Further, since each element of a column of $\B{Z}$ is independent, (\ref{zIndep2}) follows. \endproof

\section{Extended Experiments}
\paragraph{Verifying Deterministic Formulations} We verify the correctness of the deterministic formulations for various dropout schemes analyzed in this paper, i.e. \eqref{DropBlockDet} and \eqref{dropconnectNNdet}, in the top panels of Figure \ref{fig:dropblockappendix} and Figure \ref{fig:dropconnectappendix}. In Figure \ref{fig:dropblockappendix}, the curve labelled \emph{DropBlock Stochastic} is the training objective plot, \ie it plots the value of the DropBlock stochastic objective \eqref{eq:DropBlockNN} as the training progresses via Algorithm \ref{alg:sgd}. For generating the curve labeled \emph{DropBlock Deterministic}, we take the current iterate, \ie \ $\U_i, \V_i$, and plot the Deterministic  DropBlock objective obtained in Lemma \ref{lem:DropBlockDet} at every iteration. The deterministic equivalent of the DropConnect objective is similarly verified in Figure \ref{fig:dropconnectappendix}. It can be seen that the expected value of DropConnect and DropBlock over iterations matches the values derived in our results. Additionally, the bottom panel of Figure \ref{fig:dropconnectappendix} shows that Dropout and DropConnect have the same expected value of the objective at each iteration.

\begin{figure}[t]
\centering
\begin{tabular}{ccc}
$\theta=0.2$ & $\theta=0.5$ & $\theta=0.8$
\\
\hspace*{-10pt}
\includegraphics[width=0.3\textwidth]{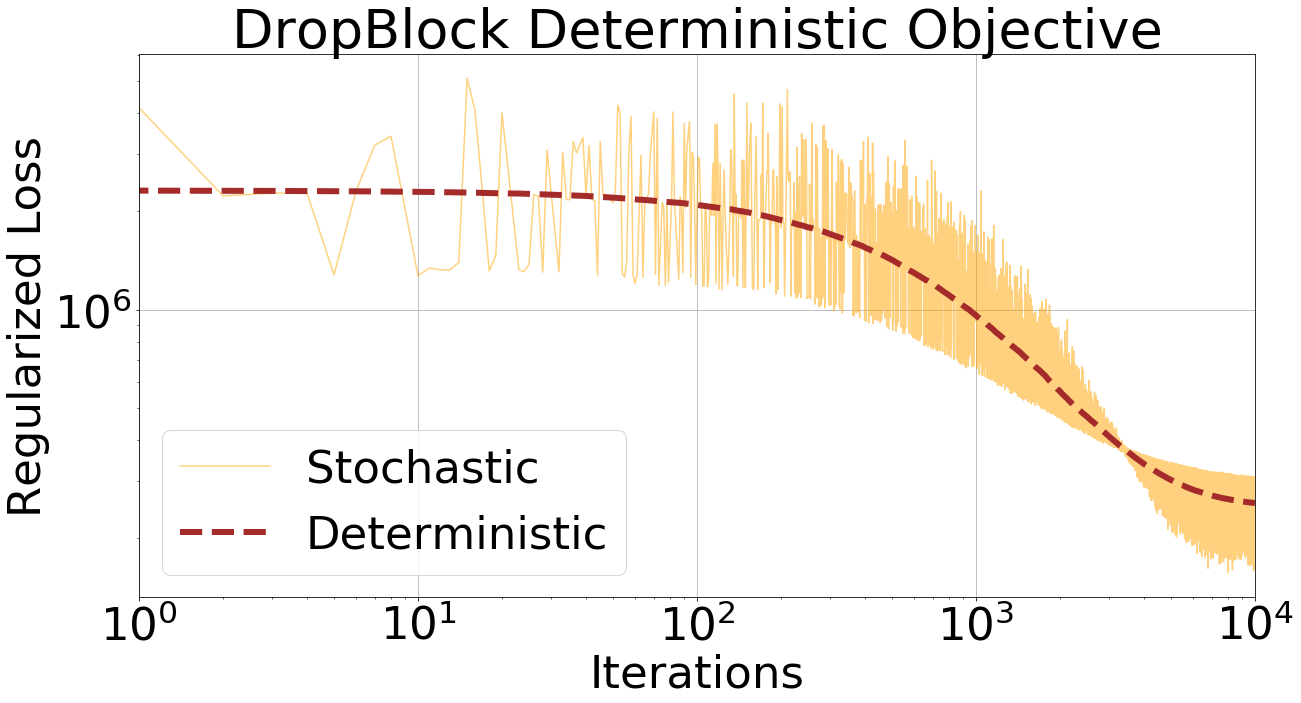}
&
\hspace*{-10pt}
\includegraphics[width=0.3\textwidth]{./fig/dropblockdet0-5.png}
&
\hspace*{-10pt}
\includegraphics[width=0.3\textwidth]{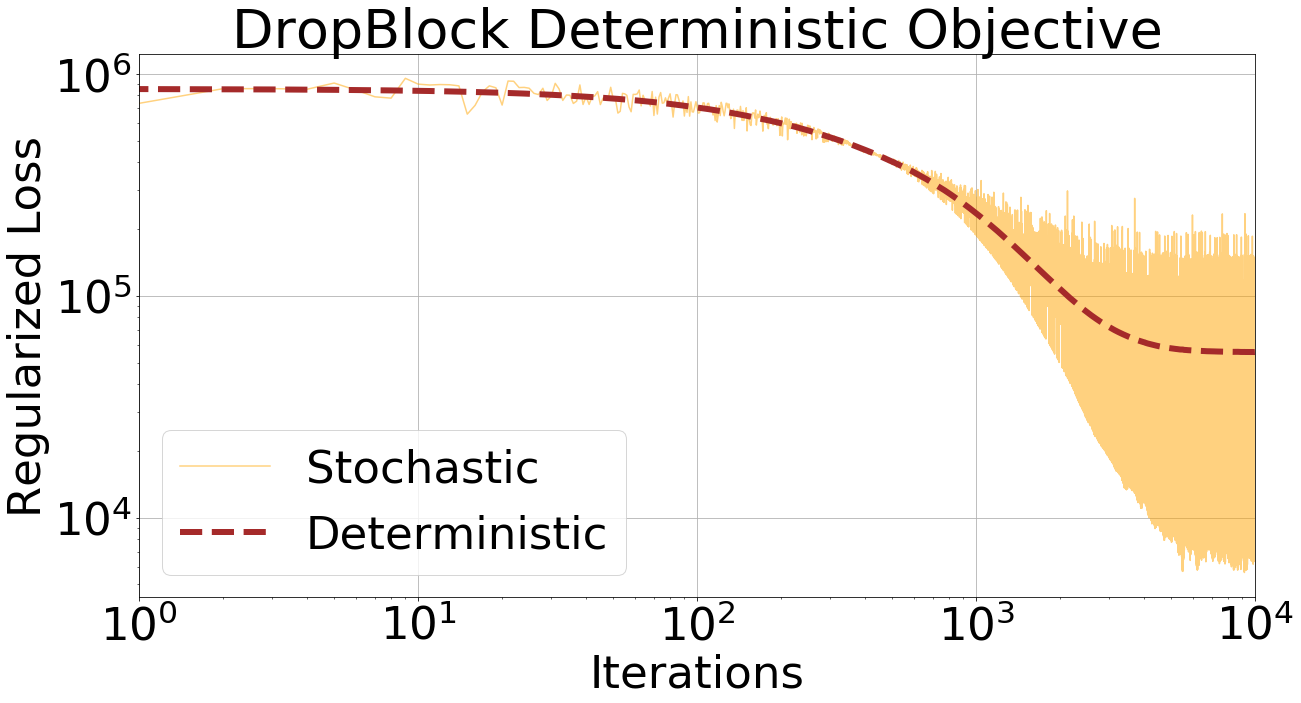} \\

\hspace*{-10pt}
\includegraphics[width=0.3\textwidth]{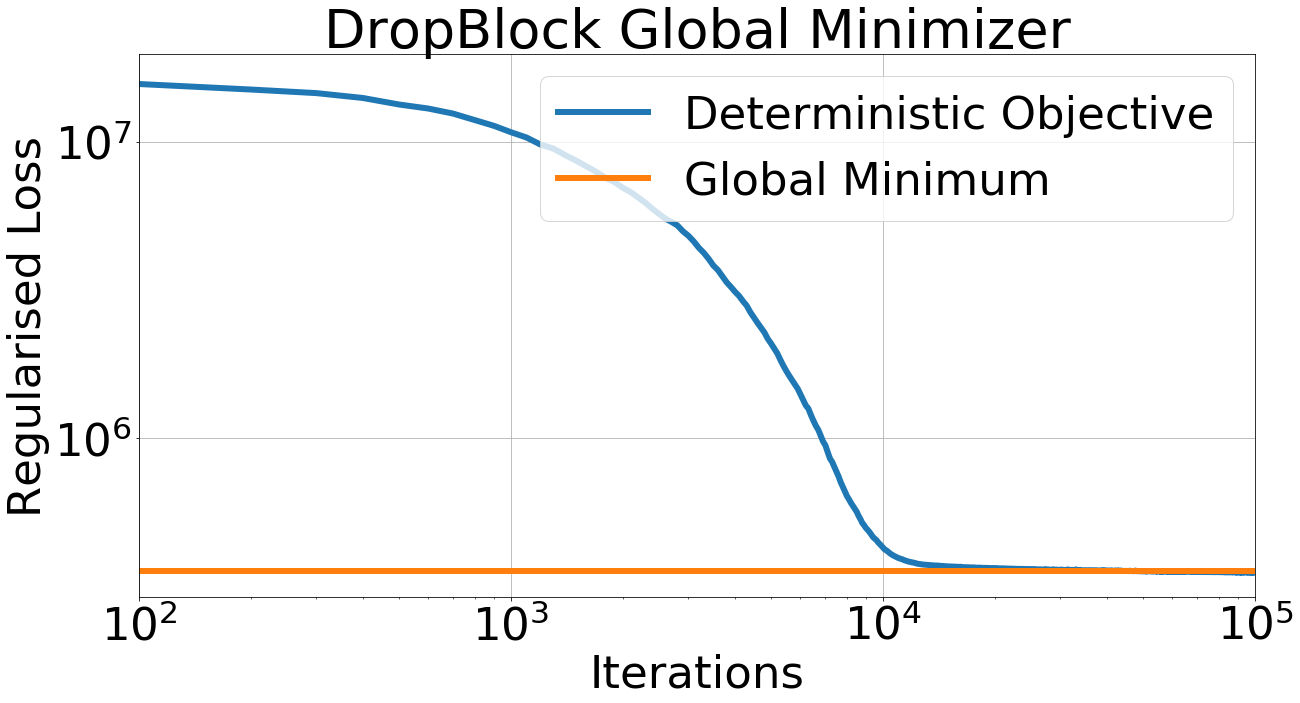}
&
\hspace*{-10pt}
\includegraphics[width=0.3\textwidth]{./fig/globmin0-5.png}
&
\hspace*{-10pt}
\includegraphics[width=0.3\textwidth]{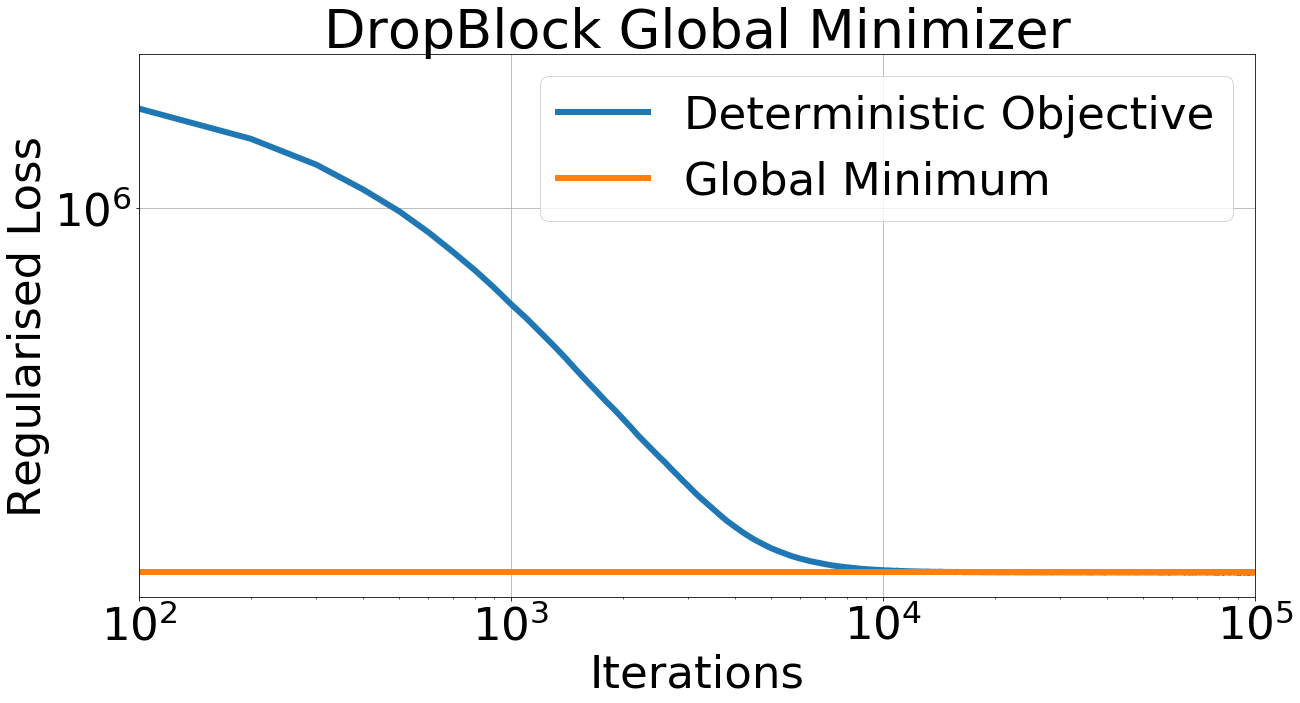}
\end{tabular}
\centering
\caption{\label{fig:dropblockappendix} Top: Stochastic DropBlockGrid training with SGD is equivalent to the deterministic Objective \eqref{DropBlockDet}. Bottom: DropBlockGrid converges to the global minimum computed in Theorem \ref{th:globmin}.}
\end{figure}

\begin{figure}[t]
\centering
\begin{tabular}{ccc}
$\theta=0.2$ & $\theta=0.5$ & $\theta=0.8$
\\
\hspace*{-10pt}
\includegraphics[width=0.33\textwidth]{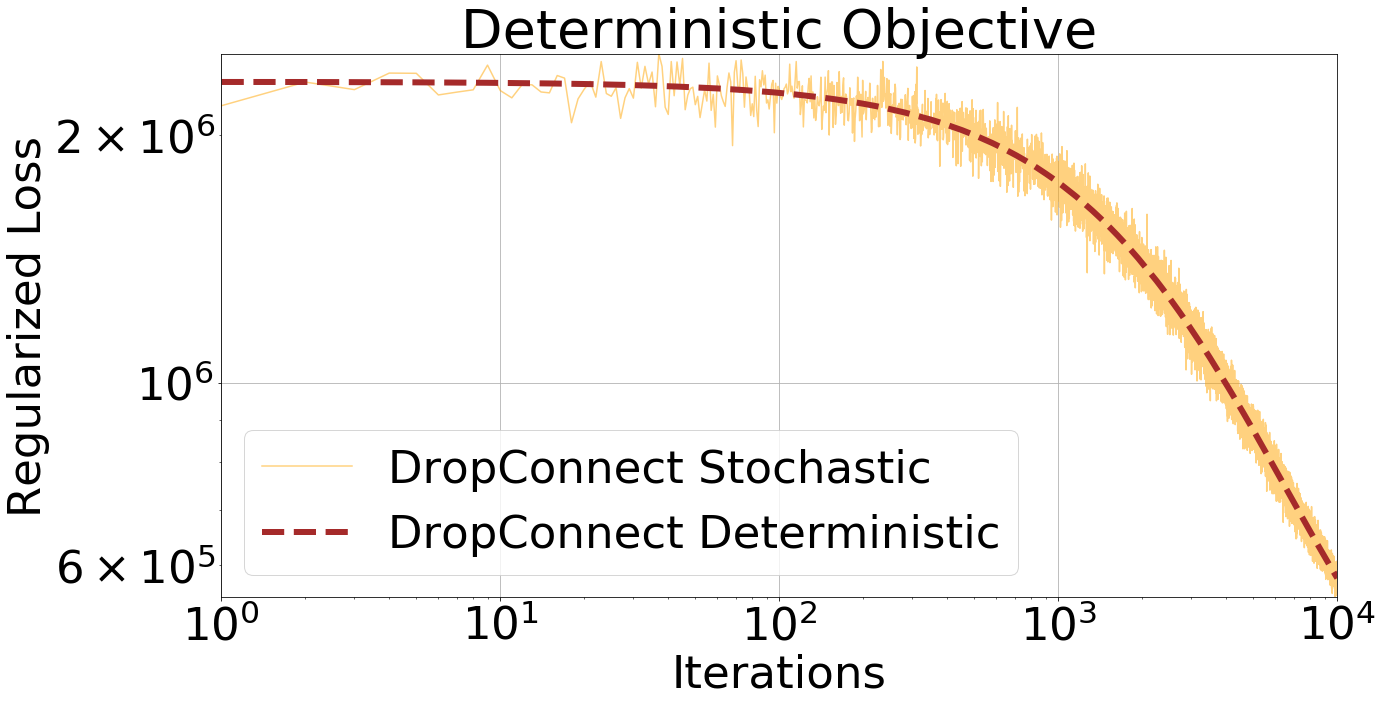}
&
\hspace*{-10pt}
\includegraphics[width=0.33\textwidth]{./fig/det0-5.png}
&
\hspace*{-10pt}
\includegraphics[width=0.33\textwidth]{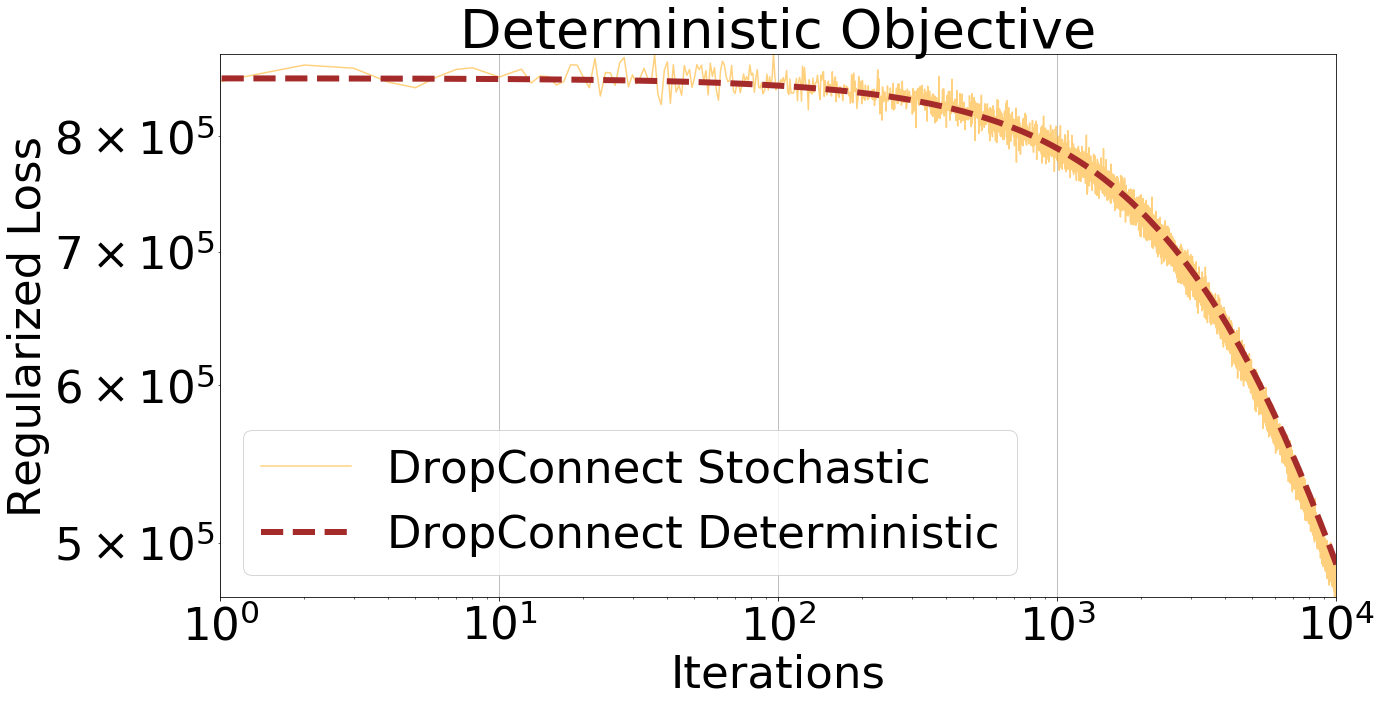} \\

\hspace*{-10pt}
\includegraphics[width=0.33\textwidth]{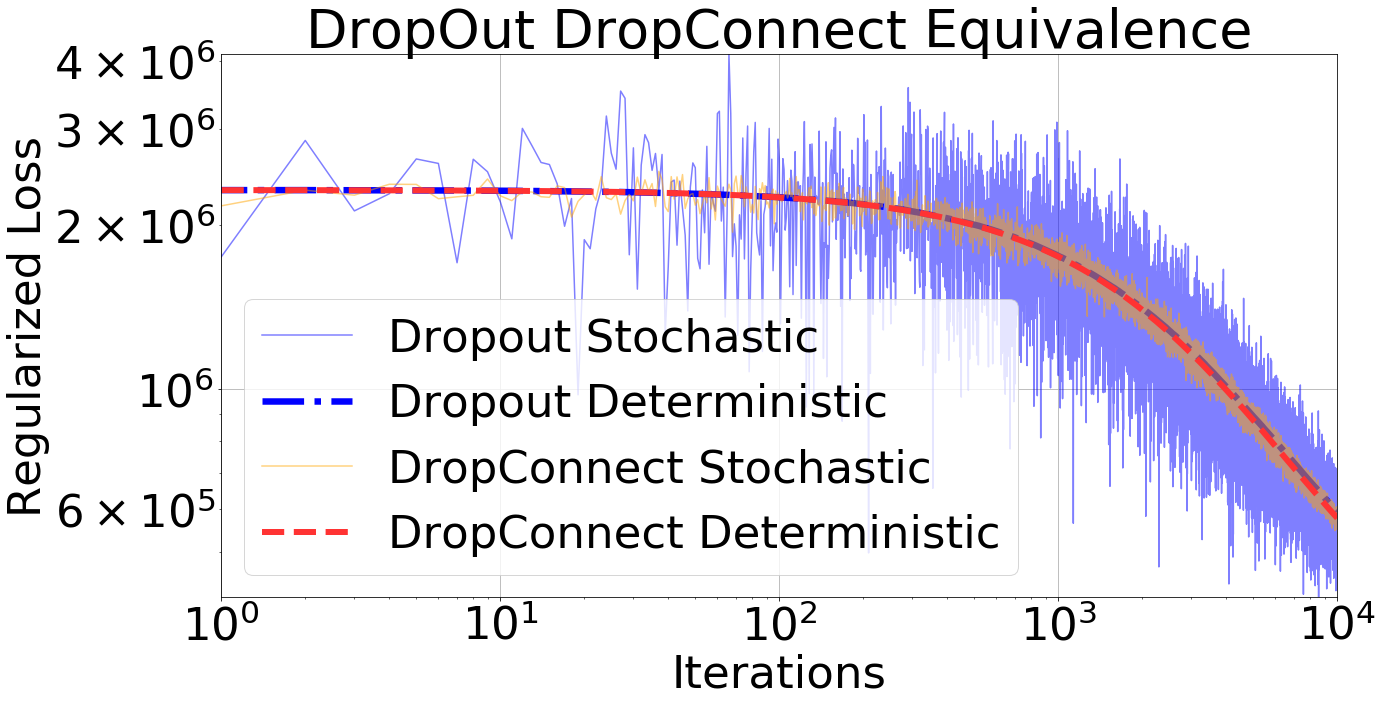}
&
\hspace*{-10pt}
\includegraphics[width=0.33\textwidth]{./fig/eqv0-5.png}
&
\hspace*{-10pt}
\includegraphics[width=0.33\textwidth]{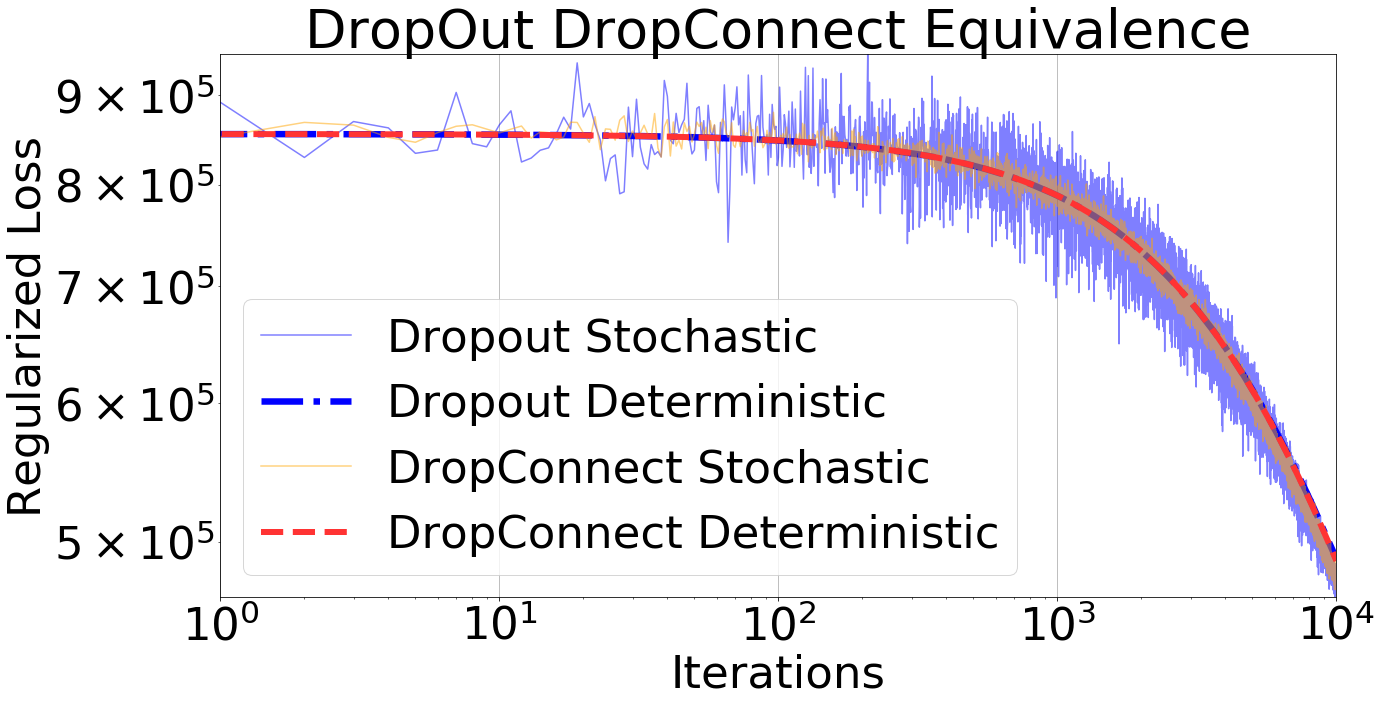}
\end{tabular}
\centering
\caption{\label{fig:dropconnectappendix} Comparing DropConnect to DropOut. Top: Stochastic DropConnect training with SGD is equivalent to the deterministic Objective \eqref{dropconnectNNdet}. Bottom: DropConnect training is equivalent to Dropout training for the squared loss.}
\end{figure}


\end{document}